\documentclass{bmvc2k}
\title{MSPred: Video Prediction at Multiple Spatio-Temporal Scales with	Hierarchical Recurrent Networks}

\addauthor{Angel Villar-Corrales$\dagger$}{villar@ais.uni-bonn.de}{1}
\addauthor{Ani Karapetyan$\dagger$}{s6ankara@uni-bonn.de}{1}
\addauthor{Andreas Boltres}{andreas.boltres@partner.kit.edu}{2, 3}
\addauthor{Sven Behnke}{behnke@cs.uni-bonn.de}{1}
\addinstitution{
	Autonomous Intelligent Systems\\
	University of Bonn\\
	Endenicher Allee 19 A,\\
	53115 Bonn, Germany
}
\addinstitution{
	SAP SE - Walldorf \\
	Dietmar-Hopp-Allee 16 \\
	69190 Walldorf, Germany
}
\addinstitution{
	Autonomous Learning Robots \\
	Karlsruhe Institute of Technology (KIT) \\
	Adenauerring 4 \\
	76131 Karlsruhe, Germany \\
}

\runninghead{Villar-Corrales et al.}{MSPred: Video Prediction at Multiple Scales}

\def\etal{\emph{et al}\bmvaOneDot}

\usepackage{amssymb}
\usepackage{amsmath}
\usepackage{algorithm}% http://ctan.org/pkg/algorithms
\usepackage[noend]{algpseudocode}
\usepackage{nth}
\usepackage[all]{nowidow}
\usepackage{bbm}
\usepackage{booktabs,etoolbox}
\usepackage{graphicx}
\usepackage{multirow}

\makeatletter
\def\algbackskip{\hskip-\ALG@thistlm}
\makeatother

\usepackage{array}
\newcolumntype{P}[1]{>{\centering\arraybackslash}p{#1}}

\newcommand{\ContextImageSet}{\mathcal{C}}

\newcommand{\PredImageSet}{\hat{\mathcal{I}}}
\newcommand{\ContextImage}{\textbf{C}}
\newcommand{\Image}{\textbf{I}}
\newcommand{\PredImage}{\hat{\textbf{I}}}

\newcommand{\PredHeatmapsSet}{\hat{\mathcal{H}}^{1}}
\newcommand{\Heatmaps}{\textbf{H}^{1}}
\newcommand{\PredHeatmaps}{\hat{\textbf{H}}^{1}}

\newcommand{\PredPositionsSet}{\hat{\mathcal{H}}^{2}}
\newcommand{\Positions}{\textbf{H}^{2}}
\newcommand{\PredPositions}{\hat{\textbf{H}}^{2}}

\newcommand{\Context}{C}
\newcommand{\NumPreds}{N}

\newcommand{\Period}{T}
\newcommand{\Latent}{\textbf{z}}
\newcommand{\Posterior}{q_{\phi}}
\newcommand{\Prior}{p_{\psi}}
\newcommand{\PrevImgs}{\Image_{1:t-1}}
\newcommand{\AllImgs}{\Image_{1:t}}

\newcommand{\Loss}{\mathcal{L}}

\newcommand{\KLDiv}{D_{KL}}

\usepackage{hyperref}
\usepackage{booktabs}                        % for nicer looking tables
\usepackage{tabularx}
\usepackage{multicol}
\newcommand{\Figure}[1]{Figure~\ref{#1}}

\newcommand{\Figures}[2]{Figures~\ref{#1} and \ref{#2}~}
\newcommand{\Figuress}[2]{Figures~\ref{#1}--\ref{#2}}

\newcommand{\Table}[1]{Table~\ref{#1}}
\newcommand{\Tables}[2]{Tables~\ref{#1}~and~\ref{#2}}
\newcommand{\Section}[1]{Section~\ref{#1}}

\let\svthefootnote\thefootnote
\newcommand\freefootnote[1]{%
	\let\thefootnote\relax%
	\footnotetext{#1}%
	\let\thefootnote\svthefootnote%
}
\usepackage{graphicscache}

\begin{document}

\maketitle
\freefootnote{$\dagger$ denotes equal contribution.}

%%%%%%%%% ABSTRACT
\vspace{-0.5cm}
\begin{abstract}
	\vspace{-0.1cm}
	Autonomous systems not only need to understand their current environment, but should also
	be able to predict future actions conditioned on past states, for instance based on captured camera frames.
	However, existing models mainly focus on forecasting future video frames for short time-horizons, hence being of limited use for long-term action planning.
	We propose \emph{Multi-Scale Hierarchical Prediction} (MSPred), a novel video prediction model able
	to simultaneously forecast future possible outcomes of different levels of
	granularity at different spatio-temporal scales. By combining spatial and temporal downsampling, MSPred efficiently predicts abstract representations such as human poses or locations over long time horizons, while still maintaining a competitive performance for video frame prediction.
	In our experiments, we demonstrate that MSPred accurately predicts future video frames as well as high-level representations (e.g. keypoints or semantics) on bin-picking and action recognition datasets, while consistently outperforming popular approaches for future frame prediction.
	Furthermore, we ablate different modules and design choices in MSPred, experimentally validating that combining features of different spatial and temporal granularity leads to a superior performance.
	Code and models to reproduce our experiments can be found in~\url{https://github.com/AIS-Bonn/MSPred}.
\end{abstract}

%%%%%%%%%%%%%%%%%%%%%%%%%%%%%%%%%%%%%%%%%%%%%%%%%%%%%%%%%%%%%%%%%%%%%%%%%%%%%%%%
\vspace{-0.3cm}
\section{Introduction}
\vspace{-0.3cm}

For effective human-robot collaboration, autonomous systems, such as domestic robots, need not only to perceive and understand their surroundings, but should also be able to estimate the intentions of nearby agents and make predictions about their actions and behavior.
Depending on the desired prediction time-horizon, the level of abstraction of the predicted representations might differ.
On the one hand, when forecasting the immediate future, predictions of high level of detail, such as subsequent video frames, are desirable.
On the other hand, for longer time horizons it is no longer possible to foresee exact details, hence it can be advantageous to predict more abstract representations like human poses or scene semantics.

Recently, several deep-learning-based methods~\cite{Denton_StochasticVideoGenerationWithALearnedPrior_2018, Saxena_ClockworkVariationalAutoencoders_2021, Villegas_HighFidelityVideoPrediction_2019, Castrejon_ImprovedVRNNsForVideoPrediction_2019, Oprea_VideoPredictionReview_2020} have been proposed for video prediction.
These methods predict future frames in an autoregressive manner, conditioned on observed or generated images, often achieving realistic predictions for short time horizons or deterministic datasets.

Despite these successes, existing models are explicitly designed to predict future video frames, either in a self-supervised manner or in a supervised setting, thus lacking the flexibility to simultaneously make predictions of different abstraction level.
Furthermore, these methods operate in an autoregressive manner in the image space, thus suffering from the increasing difficulty of predicting image-level details for longer time-horizons.
%but prediction of image-level details is increasingly more difficult for longer time-horizons.

To overcome these issues, we propose \emph{Multi-Scale Hierarchical Prediction} (MSPred), a convolutional and recurrent neural network designed to simultaneously predict future possible outcomes of different levels of abstraction with different spatial and temporal granularity.

To better model the scene dynamics and allow for better temporal modeling, MSPred utilizes a hierarchical predictor module, which applies both spatial and temporal downsampling in order to allow MSPred to extract features of different levels of abstraction that change at different temporal resolutions.
The hierarchical predictor module is composed of multiple long short-term memory~\cite{Hochreiter_LongShortTermMemory_1997} (LSTM) cells operating at different periods ($\Period$), i.e., processing every $\Period$-th input frame. LSTMs operating at a higher frequency specialize on modeling low-level fast-changing features, whereas LSTMs with a lower frequency capture more abstract representations that change at slower rates. Using this coarser temporal resolution allows us to predict abstract features far into the future, while requiring only few recurrent iterations.

In summary, our contributions are:
\textbf{(1)}~We propose \emph{MSPred}, a hierarchical video prediction model able to simultaneously predict future possible outcomes of distinct levels of abstraction at different spatio-temporal granularity, conditioned on past video frames.
\textbf{(2)}~MSPred outperforms popular video prediction models on perceptual metrics on three diverse datasets, while achieving more realistic predictions.
\textbf{(3)}~MSPred predicts plausible high-level representations (e.g. semantics or poses) long into the future using a coarse temporal resolution, while maintaining a competitive performance for video frame prediction.

%%%%%%%%%%%%%%%%%%%%%%%%%%%%%%%%%%%%%%%%%%%%%%%%%%%%%%%%%%%%%%%%%%%%%%%%%%%%%%%%
\vspace{-0.2cm}
\section{Related Work}

\subsection{Future Frame Video Prediction}

Video prediction is the task of forecasting future video frames conditioned on past frames.
This task gained popularity due to its relevance for generating anticipative behavior.
For a comprehensive review of deep-learning-based video prediction, we refer to Oprea~\etal~\cite{Oprea_VideoPredictionReview_2020}.

Several approaches have been proposed to perform video prediction, including modeling  geometric transformations between consecutive frames~\cite{Michalski_ModelingDeepTemporalDependenciesPGP_2014, Jia_DynamicFilterNetworks_2016}, learning transformations in the frequency domain~\cite{Farazi_MotionSegmentationUsingFrequencyDomainTransformerNetworks_2020,Farazi_LocalFrequencyDomainTransformerNwtworksForVideoPrediction_2021}, or using optical flow~\cite{Patraucean_SpatioTemporalVideoAutoencoderWithDifferentiableMemory_2016, Li_FlowGroundedSpatialTemporalVideoPredictionFromStillImages_2018}.
The most popular approach to video prediction, also followed by our proposed method, is the use of recurrent networks in combination with convolutional autoencoders in order to extract features from the seed frames and projecting them into the future~\cite{Srivastava_UnsupervisedLeaerningOfVideoRepresentationsUsingLSTMs_2015, Shi_ConvLSTMNetworkPrecipitationNowcasting_2015, Wang_PredRNN_2017, Wang_PredRNN_2021, Shi_DeepLearningForPrecipitationNowcastingBenchamrkAndModel_2017, Guen_DisentanglingPhysiscalDynamicsFromUnknownFactorsForVideoPrediction_2020, Akan_SlampStochasticLatentAppearanceAndMotionPrediction_2021}.
These approaches were later extended by integrating variational inference into the models, thus allowing the neural networks to model the underlying uncertainty of the data.
Stochastic prediction models~\cite{Babaeizadeh_StochasticVariationalVideoPrediction_2018,Denton_StochasticVideoGenerationWithALearnedPrior_2018,Castrejon_ImprovedVRNNsForVideoPrediction_2019} use an inference network to approximate the true posterior distribution over a set of latent variables, which model the stochastic properties of the data.
Some methods, including MSPred, organize the network into a hierarchical structure~\cite{Premont_RecurrentLadderNetworks_2017,Castrejon_ImprovedVRNNsForVideoPrediction_2019, Wu_GreedyHierarchicalVariationalAutoencoders_2021}, allowing the models to use features of different abstraction levels in order to predict future video frames.

The work that is conceptually most similar to ours is CW-VAE~\cite{Saxena_ClockworkVariationalAutoencoders_2021}, in which -- similarly to our MSPred model -- a hierarchy of recurrent modules ticking at different clock rates is used to predict higher-level features using coarse time resolutions.
Despite the similarities, the problems addressed by CW-VAE and MSPred are inherently different:
CW-VAE combines features of distinct temporal resolutions to predict detailed frames far into the future.
In contrast, we do not aim to forecast realistic frames long into the future, but instead to simultaneously predict detailed frames for short time horizons, as well as higher-level representations longer into the future using coarser temporal resolutions, which suffices for coarse-to-fine behavior planning in the now.

\subsection{High-Level Feature Prediction}

Another line of work performs video prediction using an intermediate high-level representation, instead of directly predicting future frames in the pixel space.
These models first extract some high-level representation from the seed frames, either using pretrained models or human annotations, and project them into the future. Then, the model combines the predicted structured representations and the seed frames in order to forecast the future video frames.
This approach simplifies the task of prediction, often leading to long-term accurate predictions, and has been proven successful for different high-level representations, including human poses~\cite{Villegas_LearningToGenerateLongTermFutureViaHierarchicalPredcition_2017,Fushishita_LongTermVideoGenerationOfMultipleHumanPoses_2020}, semantic segmentation maps~\cite{Pan_VideoGenerationFromSingleSemanticLabp_2019}, or instance segmentations~\cite{Luc_PredictingFutureInstanceSegmentation_2018predicting}.

Similarly, MSPred also predicts high-level representations. However, instead of simply using them as intermediate features for future frame prediction, MSPred exploits the high-level of abstraction of these representations in order to forecast them long into the future using RNNs operating with coarse temporal resolutions.

\vspace{-0.3cm}
\subsection{Multi-Scale Recurrent Networks}

Since the introduction of recurrent neural networks~\cite{Elman_FindingStructureInTime_1990,Hochreiter_LongShortTermMemory_1997} (RNNs), several approaches have been proposed to extend recurrent models into temporal hierarchies.
Hihi and Bengio~\cite{Hihi_HierarchicalRNNsForLongTermDependencies_1995} propose different architectures utilizing several RNNs operating at different time scales in order to learn long-term dependencies on simple sequential tasks.
\emph{Clockwork RNNs}~\cite{Koutnik_ClockworkRNN_2014} split a recurrent network into parallel recurrent sub-modules that process their inputs at a different temporal granularity, hence allowing the model to learn complex dependencies between temporally distant inputs.
Similarly, HM-RNN~\cite{Chung_HierarchicalMultiscaleRNNs_2016} proposes a multi-scale recurrent model with different modules operating at distinct time periods. However, the specific values for these rates are not fixed, but learned via an adaptive mechanism.

Like the previous methods, MSPred uses recurrent models operating at distinct time-scales to capture representations at different temporal resolutions. However, whereas previous methods processed low-dimensional sequences, we employ convolutional LSTMs~\cite{Shi_ConvLSTMNetworkPrecipitationNowcasting_2015} in combination with convolutional autoencoders to forecast high-dimensional video sequences.

%%%%%%%%%%%%%%%%%%%%%%%%%%%%%%%%%%%%%%%%%%%%%%%%%%%%%%%%%%%%%%%%%%%%%%%%%%%%%%%%
\vspace{-0.3cm}
\section{Method}

Video prediction is defined as the task of predicting subsequent video frames $\PredImageSet = \PredImage_{1}, \PredImage_{2}, ..., \PredImage_{\NumPreds}$ conditioned on $\Context$ past context frames $\ContextImageSet = \ContextImage_1, \ContextImage_2, ..., \ContextImage_{\Context}$.
In this work, we extend the task of video prediction to predict not only the future frames, but also higher-level representations ($\PredHeatmapsSet$, $\PredPositionsSet$), such as human poses or object locations, conditioned on the same seed frames.

\begin{figure*}[t]
	\begin{center}
		\centering
		\includegraphics[width=0.99\linewidth]{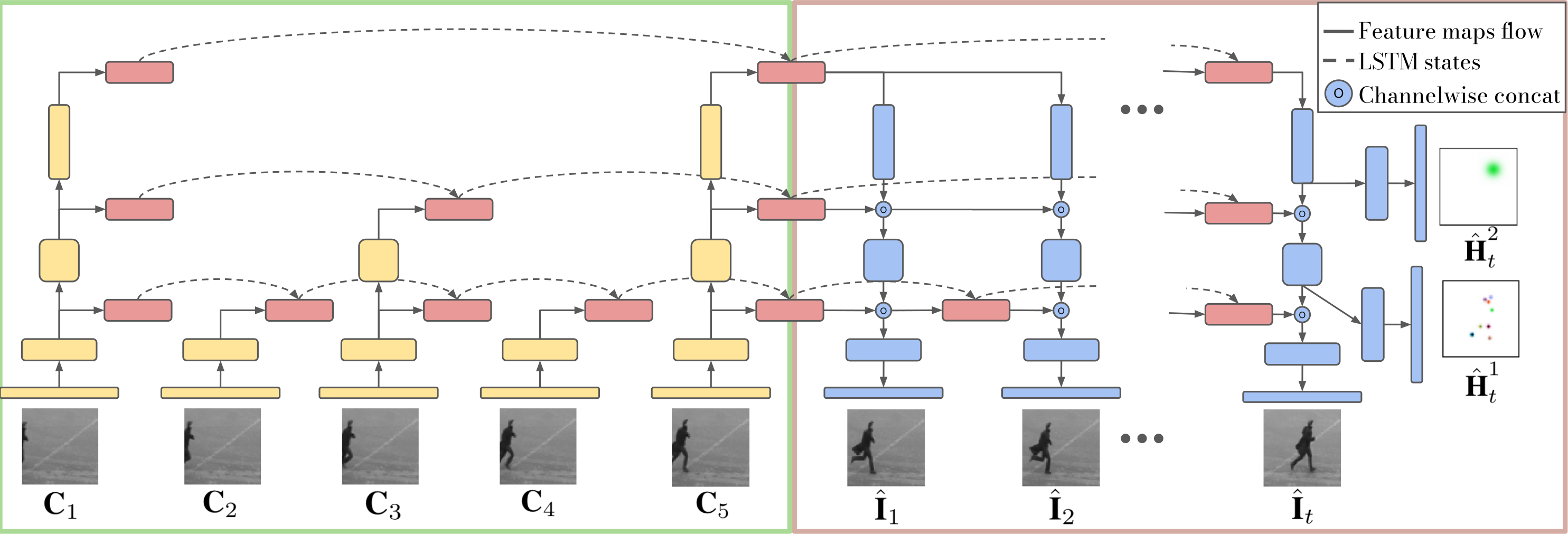}
		\vspace{-0.2cm}
		\caption{
			MSPred model.
			\textbf{Left}: Seed video frames are encoded with a {\color{Dandelion} \textbf{convolutional encoder}}.
			{\color{Salmon} \textbf{LSTM modules}} ticking at different periods forecast features at future time steps. Slower LSTMs model abstract representations, whereas the fastest LSTM processes low-level details.
			\textbf{Right}: Predicted feature maps are fed through a {\color{RoyalBlue} \textbf{decoder}}, which processes and fuses information from different LSTM modules.
			Finally, three distinct {\color{RoyalBlue} \textbf{decoder heads}} predict outputs of different abstraction level.
			Note that we do not show the stochastic modules, and only display high-level decoder heads at the last time-step to unclutter the visualization.
		}
		\label{fig:model}
	\end{center}
	\vspace{-0.5cm}
\end{figure*}

In this section we present MSPred, our proposed model for simultaneous prediction of representations of different levels of abstraction at multiple spatio-temporal scales, as illustrated in \Figure{fig:model}.
Its key component is a hierarchical predictor module~(Sec.~\ref{section: predictor}), which forecasts features of different granularity. These features are extracted from seed frames using a convolutional encoder~(Sec.~\ref{section: encoder}), and decoded into future frames or higher-level representations using convolutional decoders~(Sec.~\ref{section: decoder}).
Additionally, we discuss the stochastic components of MSPred~(Sec.~\ref{section: Stochastic}), as well as training and implementation details~(Sec.~\ref{section: details}).

\subsection{Encoder}
\label{section: encoder}

MSPred uses a 2D convolutional encoder to process the seed frames.  This module consists of four convolutional blocks, which extract increasingly abstract features of coarser spatial resolution.
After the second and third blocks, skip connections bridge from the encoder to the decoder through the two lowest-level RNNs respectively, providing features of different levels of abstraction to the decoder.
Supplying low-level representations of high-spatial resolution prevents the loss of information in the bottleneck layers, whereas higher-level features provide the decoder with abstract semantic information. The combination of these features allows our model to achieve feasible future predictions for increasingly complex structures.

\vspace{-0.2cm}
\subsection{Multi-Scale Prediction}
\label{section: predictor}

Information in images is often processed in a hierarchical manner using features of different spatial resolution and level of abstraction.
Similarly, the flow of information in videos can be represented in a temporal hierarchy. Higher-level features model the slowly changing information that is shared across many frames, whereas lower-level representations model faster-changing information.
To account for this temporal hierarchy, MSPred uses a predictor module composed of three recurrent neural networks operating at different temporal resolutions, i.e., processing input frames with a period of one, $\Period_1$ and $\Period_2$ respectively ($ 1 < \Period_1 < \Period_2$).
As depicted in \Figure{fig:model}, the recurrent module at a particular level receives feature maps from the respective stage of the encoder.
The lowest-level RNN, which processes all inputs, receives low-level feature maps of high spatial resolution, which correspond to fine-grained representations that quickly change between consecutive frames.
The second RNN receives feature maps of coarser spatial resolution, containing more slowly-changing representations, thus operating at a slower clock-rate of $\Period_1$.
Finally, the highest-level RNN receives abstract feature maps of even coarser spatial resolution, which contain high-level features that are shared across many video frames, hence processing just one input every $\Period_2$ time steps.

The use of this hierarchy of RNNs allows our model to disentangle the temporal information into three different flows, each modeling features varying at distinct time-scales.
Furthermore, the temporal abstraction in higher levels allows MSPred to forecast high-level features far into the future using just a small number of iterations, hence mitigating the error accumulation characteristic of autoregressive models.

\subsection{Decoder}
\label{section: decoder}

The decoder architecture corresponds to a mirrored version of the convolutional encoder, and is composed of four decoding stages.
After the first and second one, features from the skip connections are fused with the decoded feature maps via channel-wise concatenation. As shown in \Figure{fig:model}, MSPred uses three separate decoder heads in correspondence to its three levels of processing in order to predict representations of different level of abstraction.
Forecasting detailed video frames requires high-level knowledge (i.e. semantics or dynamics) as well as low-level information (i.e. texture and color). Therefore, the lowest-level decoder uses the most recent predicted features from all levels of the hierarchy to predict the subsequent video frames.
As higher levels operate at coarser time scales, predicted feature maps from higher levels are re-used until a new feature map is generated.
The mid-level decoder produces more abstract representations (e.g. poses or semantic segmentation) every $\Period_1$ time steps by processing the most recent feature maps of its own level and the level above.
Finally, the highest-level decoder generates abstract representations, such as person positions, every $\Period_2$ time steps using only the predicted features from the highest level.

\subsection{Stochastic Components}
\label{section: Stochastic}

Inspired by Denton and Fergus~\cite{Denton_StochasticVideoGenerationWithALearnedPrior_2018}, we enhance our MSPred model with a stochastic component to account for the uncertainty of real-world data and to achieve more diverse predictions.
During training, MSPred uses the current target frame $\Image_t$, as well as all previous frames $\PrevImgs$ to compute a posterior distribution $\Posterior(\Latent_{t}|\AllImgs)$ and sample a latent variable $\Latent_t$, which encodes the dynamics of the sequence.
By constraining the posterior to be close to a prior distribution $\Prior(\Latent_t|\PrevImgs)$, we enforce the model to encode the dynamics of the sequence, instead of simply condensing information from the target frame.
At inference time, MSPred combines previous frames $\PrevImgs$ and latent vectors $\Latent_{1:t}$ to predict future frames and representations.

In practice, we use RNNs to estimate the mean and covariance of Gaussian prior and posterior distributions, and use the reparameterization trick~\cite{Kingma_AutoEncodingVariationalBayes_2013} to sample latent vectors.
Due to the hierarchical structure of MSPred, we employ separate recurrent prior $\Prior$ and posterior $\Posterior$ modules for each level.
The latent vectors from each level are concatenated with the corresponding encoded features, and then fed to the respective predictor module~\cite{Sonderby_LadderVariationalAutoencoders_2016, Castrejon_ImprovedVRNNsForVideoPrediction_2019}.
For a more detailed description of stochastic components in video prediction, we refer to~\cite{Denton_StochasticVideoGenerationWithALearnedPrior_2018}.

\subsection{Training and Implementation Details}
\label{section: details}

Given a sequence of seed images, MSPred encodes these frames and feeds the embedded features to the corresponding recurrent modules (\Figure{fig:model} left).
During the prediction stage (\Figure{fig:model} right), the model forecasts future representations in an autoregressive manner in the feature space, i.e., the output of a recurrent module is used as input in the subsequent time step.
The forecasted features are fed to the corresponding decoder stage in order to decode future frames and high-level representations. Images are predicted at every time-step, whereas higher-level representations are predicted at the same clock-rate as their corresponding higher-level recurrent module, i.e., once every $\Period_1$ and $\Period_2$ time-steps, respectively.

For a fair comparison with baseline methods, for the Moving MNIST dataset our encoder and decoder follow DCGAN-like~\cite{Radford_DCGAN_2015} discriminator and generator architectures, respectively; whereas for other datasets we use VGG16-like~\cite{Zisserman_VGG_2014} modules.
Each level of our hierarchical predictor uses four ConvLSTM~\cite{Shi_ConvLSTMNetworkPrecipitationNowcasting_2015} cells.
The stochastic components are implemented using, for each level in the hierarchy, a single ConvLSTM cell operating at the same frequency as the corresponding predictor module.
We provide further implementation details in the supplementary material.

Similar to~\cite{Denton_StochasticVideoGenerationWithALearnedPrior_2018, Villegas_HighFidelityVideoPrediction_2019}, we include skip connections from the last observed context frame to the decoder for all prediction time-steps. The features from the skip connections are added to the outputs of the corresponding ConvLSTM.
The role of these skip connections is to directly provide the decoder with features of the background and static objects, hence allowing the predictor to focus on modeling pixel-level dynamics that change throughout the sequence.

Given $\Context$ seed frames, we train MSPred to make five predictions ($\NumPreds=5$) on each level, hence predicting the five subsequent video frames, and forecasting higher-level representations for five prediction steps with a temporal resolution of $\Period_1$ and $\Period_2$, respectively.
Our model is trained using the Adam optimizer~\cite{Kingma_Adam_2014} and the following loss function:

\vspace{-0.5cm}

\begin{align}
	& \Loss = \frac{1}{\NumPreds} \sum_{i=1}^{\NumPreds} \left( ||\Image_{i} - \PredImage_{i}||^2 + 
	\lambda_1 \Loss_{H^1}(\Heatmaps_{\Period_1 i}, \PredHeatmaps_{\Period_1 i}) + 
	\lambda_2 \Loss_{H^2}(\Positions_{\Period_2 i}, \PredPositions_{\Period_2 i}) +
	\beta \KLDiv(\Posterior, \Prior)
	\right), \nonumber
\end{align}

\vspace{-0.2cm}

where $|| \cdot ||$ is the $\ell_2$ norm, $\Image_t, \Heatmaps_t, \Positions_t$ correspond to the ground truth frames and higher-level targets at time-step $t$, $\PredImage_t, \PredHeatmaps_t, \PredPositions_t$ are the corresponding predictions of each level at time-step $t$, and $\lambda_1$, $\lambda_2$ and $\beta$ are coefficients to weight the different loss terms.
$\Loss_{H^1}$ and $\Loss_{H^2}$ correspond the loss functions used on the higher-level decoder heads. We employ the mean-squared error when predicting poses or positions, and pixel-wise cross-entropy when forecasting semantic maps.
Finally, $\KLDiv$ corresponds to the KL-divergence error between
the posterior $\Posterior$ and prior $\Prior$ distributions estimated by the stochastic component, averaged over the three levels in the hierarchy.

%%%%%%%%%%%%%%%%%%%%%%%%%%%%%%%%%%%%%%%%%%%%%%%%%%%%%%%%%%%%%%%%%%%%%%%%%%%%%%%%
\vspace{-0.2cm}
\section{Experiments}
\vspace{-0.2cm}

We perform an ablation study investigating several MSPred modules and design choices in~\Section{section: ablation}.
In~\Section{section: comparison} we compare MSPred with existing video prediction models on three diverse datasets.
Finally, in~\Section{section: mspred} we present results of multi-scale prediction.
Animations and further qualitative results are available in the project website\footnote{\url{https://sites.google.com/view/mspred/home}}.
%:\url{https://sites.google.com/view/mspred}.
%
%\href{https://sites.google.com/view/mspred/home}{project website}.

%%%%%%%%%%%%%%%%%%%%%%%%%%%%%%%%%%%%%%%%%%%%%%%%%%%%%%%%%%%%%%%%%%%%%%%%%%%%%%%%
\vspace{-0.3cm}
\subsection{Datasets}
\label{section: evaluation}

We evaluate MSPred for different prediction tasks on three video datasets of different levels of complexity, namely Moving MNIST~\cite{Srivastava_UnsupervisedLeaerningOfVideoRepresentationsUsingLSTMs_2015}, KTH-Actions~\cite{Schuldt_KTHRecognizingHumanActionsALocalSVMApproach_2004}, and SynpickVP.
In all cases, we average the results across five predicted frames or high-level representations.
Further details about datasets and evaluation metrics are provided in the supplementary material.

\textbf{Moving MNIST} is a standard video prediction dataset containing sequences of two random digits from the MNIST dataset~\cite{LeCun_MnistDataset_1998} moving with constant speed in a $64 \times 64$ grid, and bouncing off the image boundaries. We train our models on random sequences generated on the fly, and evaluate on a fix test set containing 10,000 sequences. 

\textbf{KTH-Actions} is a dataset consisting of real videos of humans performing one out of six possible actions, e.g. jogging or waving. The dataset includes 600 videos of 25 different humans performing the actions in various indoor and outdoor environments.

\textbf{SynpickVP} is a video prediction dataset containing sequences of bin-picking scenarios, in which a suction-cap gripper robot moves objects in a cluttered box.
We generate the dataset by selecting sequences from the recently proposed SynPick~\cite{Periyasamy_SynPick_2021} dataset.
This is a challenging benchmark, in which the model needs to predict the motion of the robotic gripper, as well as the displaced objects, while representing a complex and cluttered background.

\begin{table*}[t!]
	\centering
	\caption{Ablation study investigating several MSPred modules and design choices, i.e. different RNNs, temporal and spatial hierarchy, and the effect of supervision on the higher levels of the hierarchy. Best result is highlighted in boldface, second best is underlined.
	}
	\label{table: ablation study}
	\vspace{0.05cm}
	\small
	\begin{tabular}{p{0.cm} p{0.66cm} wc{0.5cm} wc{1.8cm} wc{1.85cm} p{0.1cm}p{0.9cm}p{0.9cm}p{0.9cm}p{0.9cm}}
		\toprule
		\multicolumn{5}{c}{\textbf{MSPred Modules}} && \multicolumn{4}{c}{\textbf{Video Prediction Results}} \\
		& RNN & Spatial & Temporal & Hierarch. Supervision && MSE$\downarrow$ & PSNR$\uparrow$ & SSIM$\uparrow$ & LPIPS$\downarrow$ \\
		\midrule
		1 & Conv. & \hfill \checkmark & \hfill \checkmark  & \checkmark && \textbf{41.52} & \textbf{25.99} & \textbf{0.970} & \textbf{0.030} \\
		2 & Conv.  & \hfill \checkmark & \hfill \checkmark & \hfill - && \underline{42.47} & \underline{25.94} & \textbf{0.970} & \textbf{0.030} \\
		3 & Linear & \hfill \checkmark & \hfill \checkmark & \checkmark && 208.71 & 17.95 & 0.827 & 0.202 \\
		4 & Conv. & \hfill - & \hfill  \checkmark & \checkmark && 73.47 & 22.81 & \underline{0.950} & \underline{0.057} \\
		5 & Conv. & \hfill \checkmark & \hfill -  & \checkmark && 92.45 & 20.81 & 0.921 & 0.093 \\
		6 & Conv. & \hfill - & \hfill - & \hfill - && 112.18 & 20.97 & 0.912 & 0.097 \\
		\bottomrule
	\end{tabular}
	\vspace{-0.cm}
\end{table*}

\subsection{Ablation Study}
\label{section: ablation}

We train modified variants of MSPred to investigate how different components and design choices affect the video prediction performance of our model. Namely, we evaluate the relevance of temporal and spatial hierarchy, the type of RNN used, and the effect of multi-level supervision. For our ablation study, we focus on the Moving MNIST dataset.
The results are reported in \Table{table: ablation study}.

First, the hierarchical MSPred model (Rows 1 and 2) outperforms all other variants, whereas removing the spatial (Row 4), temporal (Row 5), or all (Row 6) hierarchical structure from the predictor leads to loss of performance, thus demonstrating that providing features of different spatio-temporal granularity improves the prediction performance of the model.
Second, MSPred (Row 1) performs comparably to an MSPred variant trained only for image-level prediction (Row 2), thus indicating that adding hierarchical supervision to the loss function is not a key factor for the success of our model for future frame prediction.
Finally, when replacing the ConvLSTM predictor with a linear LSTM (Row 3), the performance is significantly decreased, leading to the worst results among all compared models.

\begin{table*}[t!]
	\centering
	\caption{Quantitative comparison between video prediction models. MSPred outperforms all other methods on Moving MNIST, and achieves the best perceptual results (LPIPS) on KTH-Actions and SynpickVP. Best result is highlighted in boldface, second best is underlined.}
	\label{table: comparison}
	\vspace{0.05cm}
	\small
	\begin{tabular}{p{1.98cm} p{0.64cm}p{0.63cm}p{0.63cm}r p{0.64cm}p{0.63cm}p{0.63cm}r  p{0.64cm}p{0.63cm}p{0.76cm}}
		\toprule
		&  \multicolumn{3}{c}{\textbf{Moving MNIST}} && \multicolumn{3}{c}{\textbf{KTH-Actions}} && \multicolumn{3}{c}{\textbf{SynpickVP}}\\
		
		\textbf{} & {PSNR}$\uparrow$ & {SSIM}$\uparrow$ & {LPIPS}$\downarrow$ && {PSNR}$\uparrow$ & {SSIM}$\uparrow$ & {LPIPS}$\downarrow$ && {PSNR}$\uparrow$ & {SSIM}$\uparrow$ & {LPIPS}$\downarrow$ \\
		\midrule
		ConvLSTM~\cite{Shi_ConvLSTMNetworkPrecipitationNowcasting_2015} & 17.22 & 0.833 & 0.144  &&  \underline{29.93} & \underline{0.957} & 0.048 && 27.98 & \underline{0.907} & 0.059	\\
		TrajGRU~\cite{Shi_DeepLearningForPrecipitationNowcastingBenchamrkAndModel_2017} & 20.02 & 0.895 & 0.075	 && \textbf{30.02} & \textbf{0.958} & 0.039 && 28.10 & \textbf{0.908} & 0.041	\\
		SVG-Det~\cite{Denton_StochasticVideoGenerationWithALearnedPrior_2018} & 20.31 & 0.900 & 0.114	 &&  26.64 & 0.927 & 0.068 && 26.92 & 0.879 & 0.068	\\
		SVG-LP~\cite{Denton_StochasticVideoGenerationWithALearnedPrior_2018} & 20.36 & 0.907 & 0.115	 && 27.60 & 0.932 & 0.063 && 27.38 & 0.886 & 0.066	\\
		PredRNN++~\cite{Wang_PredRNN_2021} & 20.20 & 0.911 & 0.055	 && 29.51 & 0.941 & 0.068 && 27.50 & 0.894 & 0.053	\\
		PhyDNet~\cite{Guen_DisentanglingPhysiscalDynamicsFromUnknownFactorsForVideoPrediction_2020} & \underline20.43 & \underline{0.915} & \underline{0.054}	 && 28.01 & 0.913 & 0.125 && 26.84 & 0.877 & 0.053	\\
		MSPred~NoSup & \underline{25.94} & \textbf{0.970} & \textbf{0.030} && 28.65 & 0.929 & \underline{0.034} && \textbf{28.92} & 0.902 & \underline{0.031}	\\
		MSPred~(ours) & \textbf{25.99} & \textbf{0.970} & \textbf{0.030} && 28.93 & 0.930 & \textbf{0.032} && \underline{28.61} & 0.903 & \textbf{0.030}	\\
		\bottomrule
	\end{tabular}
	\vspace{-0.cm}
\end{table*}

\subsection{Comparison to Existing Methods}
\label{section: comparison}

We compare MSPred for the task of future frame prediction with several existing video prediction methods, including ConvLSTM~\cite{Shi_ConvLSTMNetworkPrecipitationNowcasting_2015}, TrajectoryGRU~\cite{Shi_DeepLearningForPrecipitationNowcastingBenchamrkAndModel_2017}, two variants of SVG~\cite{Denton_StochasticVideoGenerationWithALearnedPrior_2018}, PredRNN++~\cite{Wang_PredRNN_2021}, and PhyDNet~\cite{Guen_DisentanglingPhysiscalDynamicsFromUnknownFactorsForVideoPrediction_2020}.
Additionally, we include an MSPred variant, denoted as \emph{MSPred NoSup}, trained only for future frame prediction, i.e., without any additional supervision.
For a fair comparison, all models use the same encoder and decoder architectures.
The evaluation results are reported in \Table{table: comparison}.

We observe that both MSPred variants consistently perform among the best models on all datasets. 
Furthermore, MSPred NoSup shows comparable video prediction performance to MSPred trained to simultaneously make predictions of different levels of granularity.
This indicates that adding hierarchical supervision to the loss function (e.g. poses or segmentation) allows MSPred to predict higher-level representations over
long time-horizons, but it is not a key factor for the success of our model for video prediction.

\textbf{Moving MNIST:} In general, due to the simplicity of the dataset, all models achieve overall good prediction scores and accurate future frame predictions. However, both MSPred variants achieve exceptionally sharp and precise reconstructions, outperforming all other models by a large margin.

\textbf{KTH-Actions:} MSPred produces low PSNR and SSIM scores, indicating higher pixel differences with respect to the target frames. However, MSPred achieves the best LPIPS result, demonstrating a high perceptual similarity to the target frames.
\Figure{fig:kth qual} contains a qualitative comparison on three KTH-Actions sequences.
Baseline methods with the highest PSNR results, i.e. ConvLSTM and TrajGRU, blur the predictions around the arms and legs of the person, whereas our MSPred model achieves sharper predictions.

\textbf{SynpickVP:} MSPred outperforms existing video prediction models on this challenging dataset, achieving the highest PSNR result, and the best LPIPS perceptual score, indicating more realistic predictions than the baselines.
\Figure{fig:synpick qual} depicts a qualitative comparison on the SynpickVP dataset.
Baseline video prediction methods tend to blur the suction cap gripper as well as the objects moved by it, whereas MSPred achieves more realistic predictions.

\subsection{Multi-Scale Prediction}
\label{section: mspred}

Unlike existing video prediction models, MSPred predicts higher-level representations with  coarse temporal resolutions in addition to video frames.
For KTH-Actions, MSPred predicts human keypoints on its intermediate level, and a center-point on its highest level.
Keypoint annotations are obtained using OpenPose~\cite{Cao_RealTimePoseEstimationAffinityFields_2017}.
On SynpickVP, MSPred predicts semantic segmentation and the location of the gripper on the mid- and high-levels, respectively.
We average the segmentation results in three different groups: \emph{gripper}, \emph{static}, and \emph{background}.

For benchmarking the performance of MSPred, we train a baseline model \emph{SVG'}, based on a modified SVG-LP~\cite{Denton_StochasticVideoGenerationWithALearnedPrior_2018} model, which predicts high-level representations conditioned on input frames. Further details about SVG' are provided in the supplementary material.
Furthermore, we include a CopyLast baseline that copies the ground-truth representations of the last seed frame. 

\Table{table: ms} reports a quantitative comparison for pose and semantic segmentation forecasting, respectively.
MSPred outperforms the two baselines for human pose forecasting on the KTH-Actions dataset.
On the SynpickVP dataset, MSPred precisely predicts the segmentation of the robot gripper.
However, due to the complex and mostly static scenes, the CopyLast baseline outperforms SVG' and MSPred when forecasting the segmentation of static objects and background.
This is due to the fact that CopyLast has access to the last ground-truth representations, thus having access to the perfect segmentation of static objects and background; whereas MSPred and SVG' predict the high-level representations condition on seed images, without access to ground truth segmentation maps.

\begin{figure}[t]
	\begin{tabular}{cc}
		\begin{minipage}{0.4\textwidth}
			\vspace{-0.02cm}
			\hspace{-0.28cm} \includegraphics[width=1\linewidth]{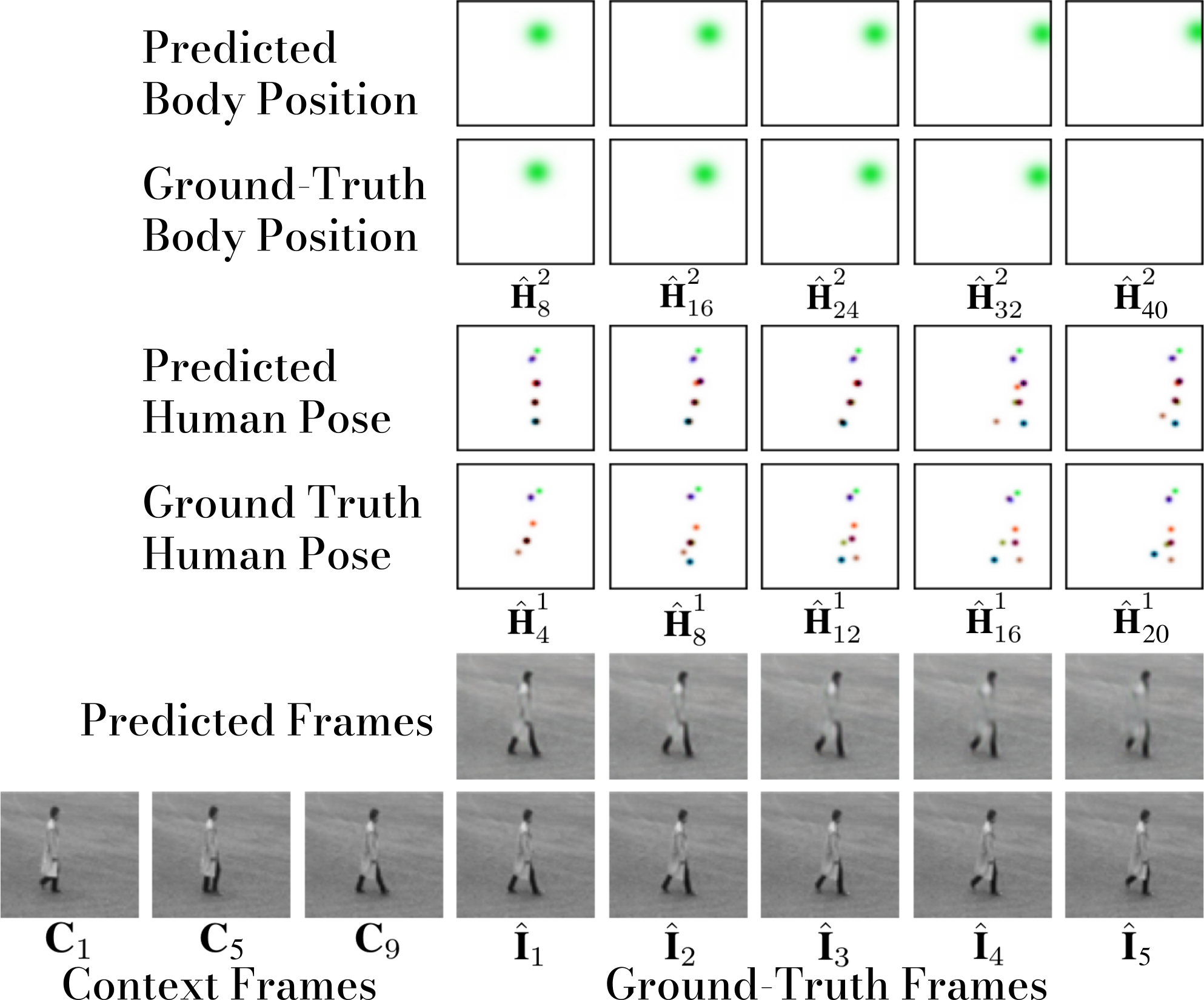} \\
		\end{minipage}
		&
		
		\begin{minipage}{0.59\textwidth}
			\vspace{-0.5cm}
			\hspace{-0.64cm} \includegraphics[width=1\linewidth]{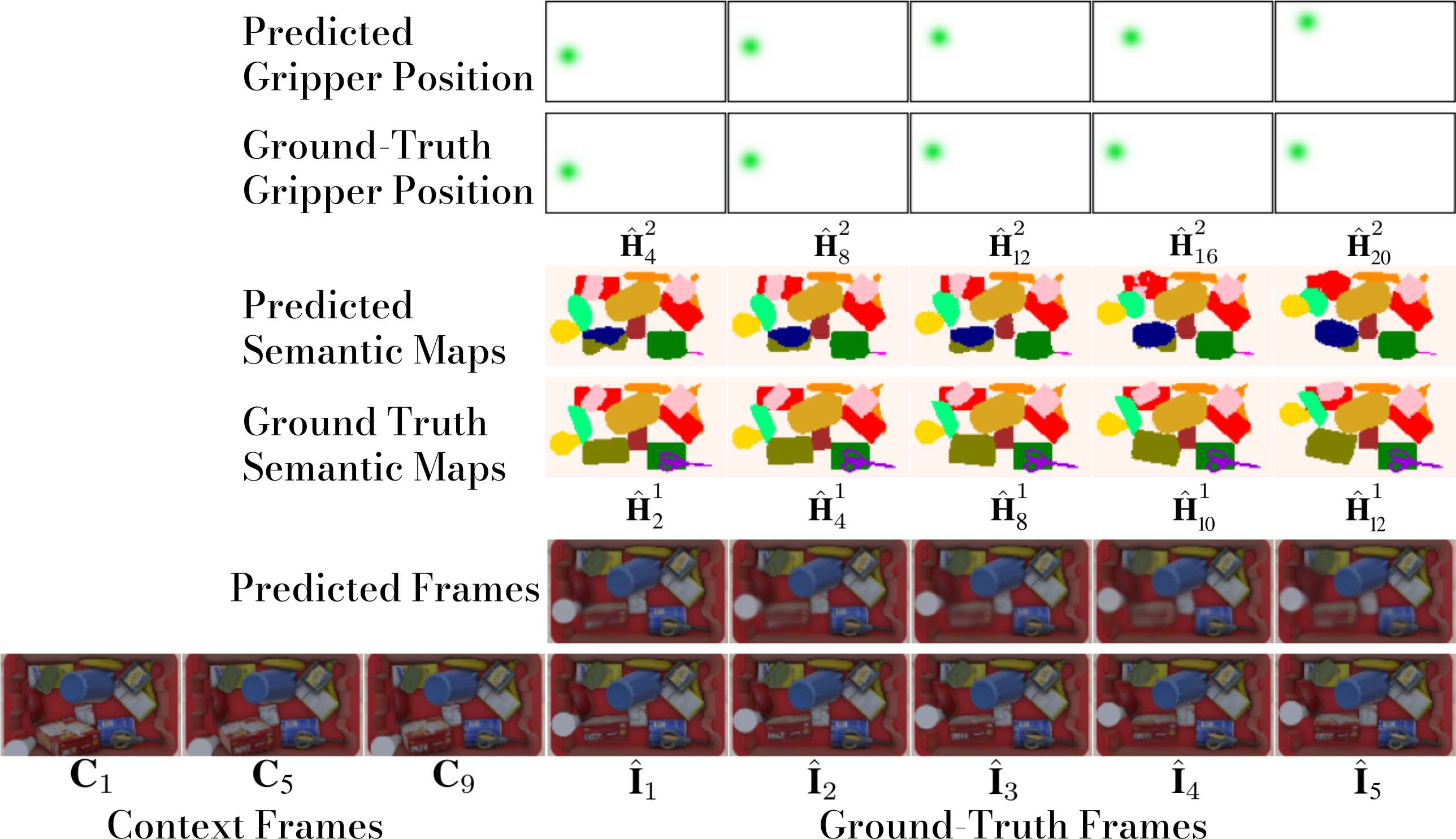}
		\end{minipage}	
	\end{tabular}
	\vspace{-0.7cm}
	\caption{
		Multi-scale predictions on the KTH-Action (left) and SynpickVP (right) datasets. We display three seed frames and five targets and predictions for each decoder head. 
		MSPred forecasts frames on short time horizons while also predicting higher-level representations longer into the future using a coarse temporal resolution.
	}
	\label{fig: ms}
	\vspace{-0.05cm}
\end{figure}

\Figure{fig: ms} depicts examples of multi-scale prediction on the KTH-Actions and SynpickVP datasets.
The lowest-level decoder achieves detailed subsequent frame predictions for a short time horizon, whereas higher level decoders accurately predict abstract representations (e.g. poses or locations) up to 40 frames into the future using coarser temporal resolutions.

\begin{figure}[t]
	\begin{tabular}{ccc}
		\begin{minipage}{0.33\textwidth}
			\hspace{-0.28cm} \includegraphics[width=1\linewidth]{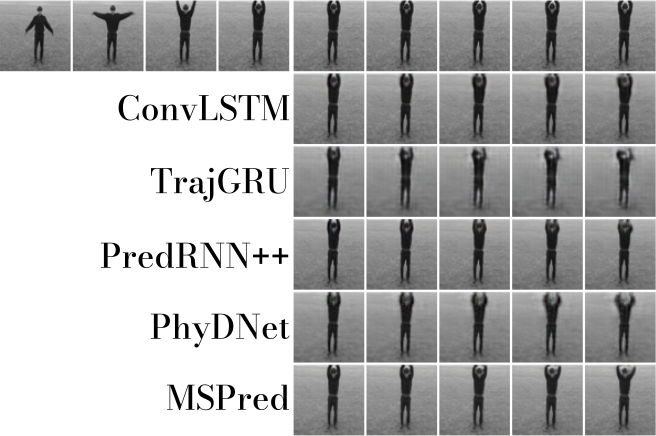} \\
		\end{minipage}
		&
		
		\begin{minipage}{0.33\textwidth}
			\vspace{-0.43cm}
			\hspace{-0.66cm} \includegraphics[width=1\linewidth]{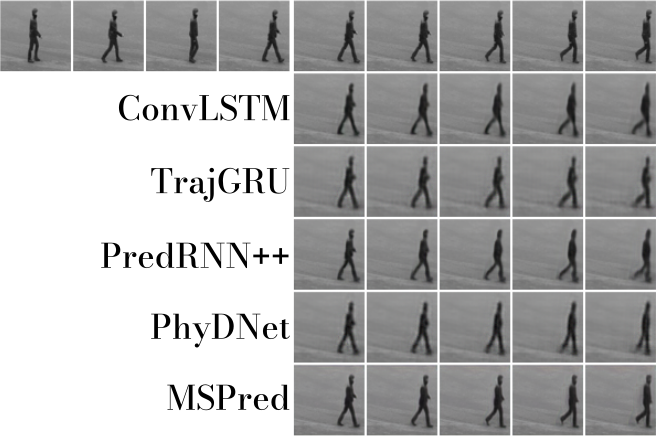}
		\end{minipage}	
		&
		\begin{minipage}{0.33\textwidth}
			\vspace{-0.43cm}
			\hspace{-1.04cm} \includegraphics[width=1\linewidth]{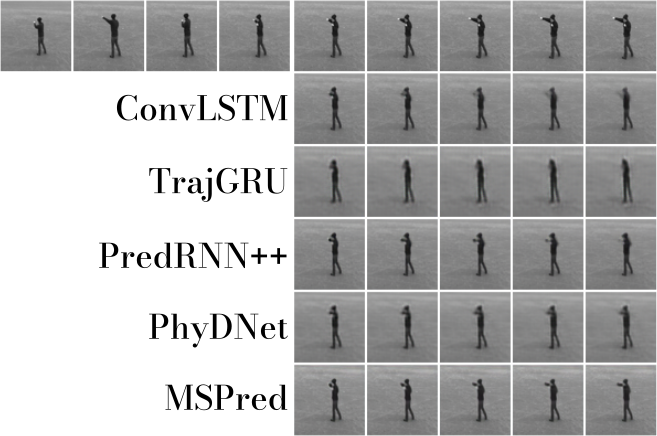}
		\end{minipage}	
	\end{tabular}
	\vspace{-0.7cm}
	\caption{
		Qualitative results on the KTH-Actions dataset. 
		Top row shows ground truth frames. 
		We display four seed frames and five predictions for three test-set sequences.
		MSPred achieves the sharpest and most accurate predictions among the compared methods.
	}
	\label{fig:kth qual}
	\vspace{-0.1cm}
\end{figure}

\begin{figure}[t]
	\vspace{-0.1cm}
	\begin{tabular}{ccc}
		\begin{minipage}{0.496\textwidth}
			\hspace{-0.28cm} \includegraphics[width=1\linewidth]{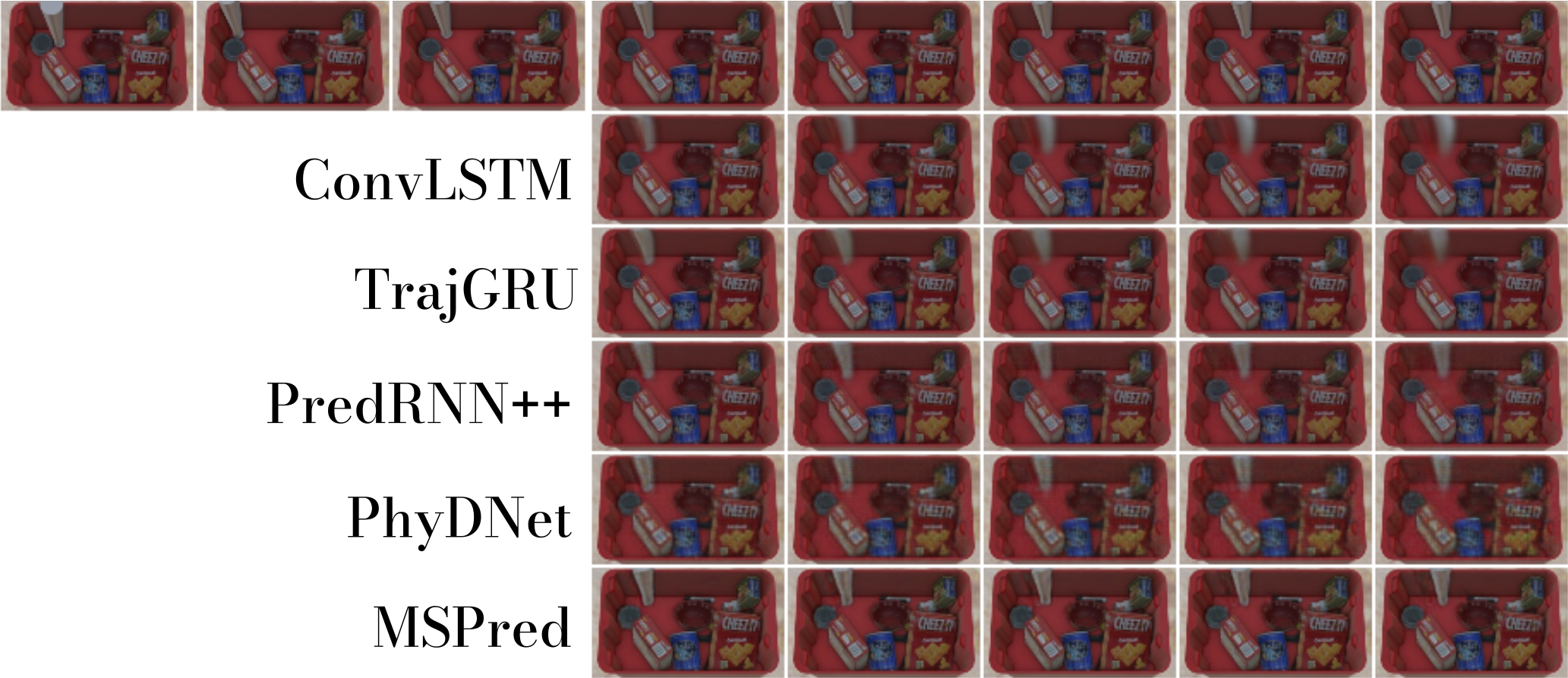} \\
		\end{minipage}
		&
		\begin{minipage}{0.496\textwidth}
			\vspace{-0.45cm}
			\hspace{-0.66cm} \includegraphics[width=1\linewidth]{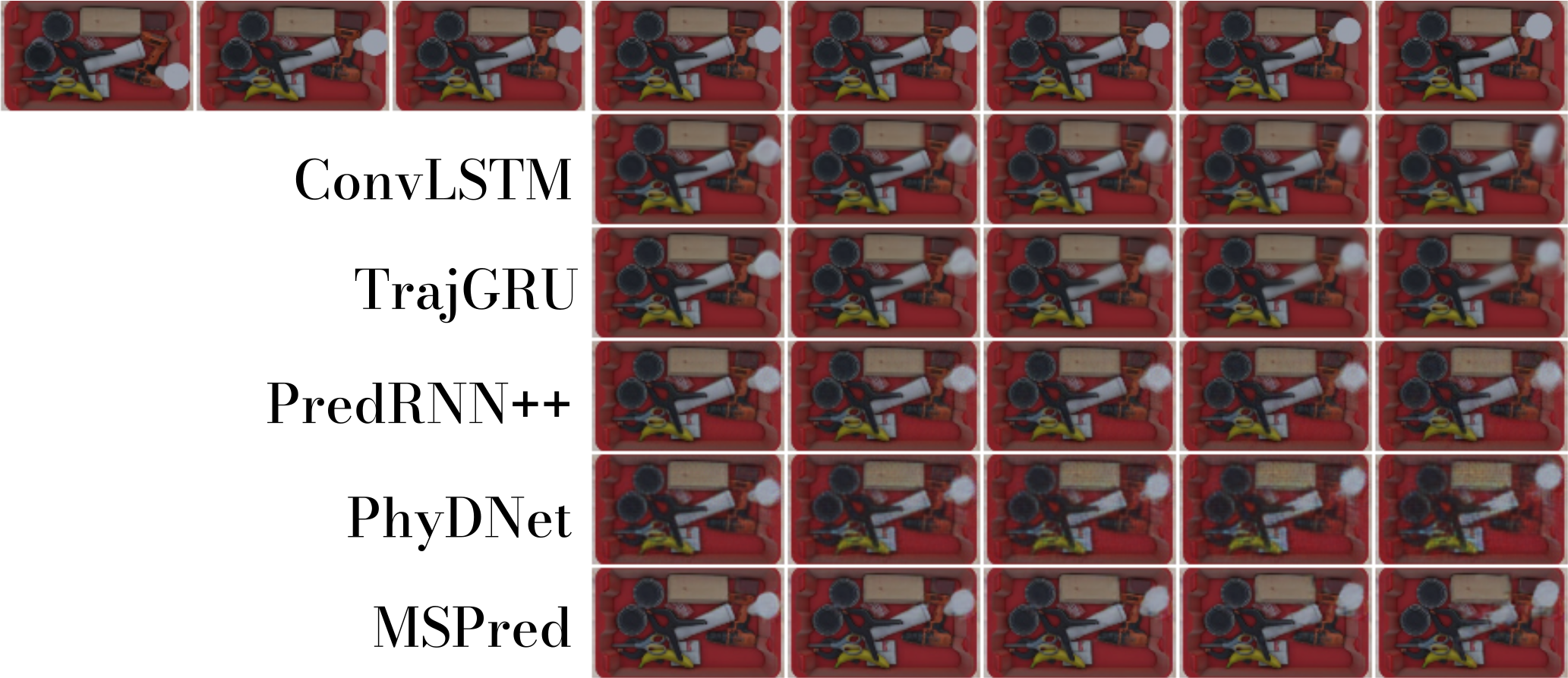}
		\end{minipage}	
	\end{tabular}
	\vspace{-0.7cm}
	\caption{
		Qualitative results on the SynpickVP dataset. 
		Top row shows ground truth frames. 
		We display three seed frames and five predictions for two test-set sequences.
		Whereas all baseline methods blur the robot gripper in the predicted frames, MSPred achieves sharper predictions,
	}
	\label{fig:synpick qual}
	\vspace{-0.4cm}
\end{figure}

%%%%%%%%%%%%%%%%%%%%%%%%%%%%%%%%%%%%%%%%%%%%%%%%%%%%%%%%%%%%%%%%%%%%%%%%%%%%%%%%
\vspace{-0.3cm}
\section{Conclusion}
\vspace{-0.3cm}

We proposed MSPred, a novel video prediction model that extends the effective prediction horizon of related approaches by leveraging hierarchies of recurrent neural networks operating at different spatio-temporal resolutions in order to predict outcomes of varying levels of abstraction with different granularity.
At its lowest prediction level, MSPred forecasts subsequent video frames, whereas at higher levels it predicts more abstract representations longer into the future using coarser spatio-temporal resolutions.
In our experiments, we show how MSPred outperforms several existing video prediction methods for the task of future frame prediction. Furthermore, we show how the higher level decoder heads can be used to forecast more abstract representations, such as human poses or semantic maps, over longer time horizons.
We firmly believe that the hierarchical features from MSPred could be used as representations to improve perception and reasoning capabilities in autonomous agents, and serve as basis for planning anticipative behavior.

\begin{table}[t]
	\caption{
		Comparison between MSPred and two baselines on pose prediction and segmentation forecasting.
		Best result is highlighted in boldface, second best is underlined.
	}
	\label{table: ms}
	\vspace{-0.cm}
	\small
	\begin{tabular}{p{1.2cm} P{0.9cm}p{0.6cm}p{0.6cm}p{0.9cm}  p{0.55cm}p{0.4cm}p{0.55cm}p{0.7cm}p{0.55cm}p{0.55cm}}
		\toprule
		&  \multicolumn{4}{c}{\textbf{KTH-Actions}} &  \multicolumn{6}{c}{\textbf{SynpickVP}} \\
		&  \multicolumn{4}{c}{\textbf{Pose Forecasting}} & \multicolumn{2}{c}{\textbf{Gripper Seg.}} & \multicolumn{2}{c}{\textbf{Static Seg.}}
		& \multicolumn{2}{c}{\textbf{Backgr. Seg.}} \\
		
		& MPJPE$\downarrow$ & PCK$\uparrow$ & PDJ$\uparrow$ & AP$\uparrow$ & Acc$\uparrow$ & IoU$\uparrow$ & Acc$\uparrow$ & IoU$\uparrow$ & Acc$\uparrow$ & IoU$\uparrow$ \\ 
		
		\midrule
		
		CopyLast & 10.50 & 0.550 & 0.558 & 0.620 & 0.397 & 0.324 & \textbf{0.557} & \textbf{0.455} & \textbf{0.957} & \textbf{0.919} \\
		
		SVG'~\cite{Denton_StochasticVideoGenerationWithALearnedPrior_2018} & \underline{4.32} & \underline{0.810} & \textbf{0.888} & \underline{0.825} & 0.606 & 0.443 & \underline{0.255} & \underline{0.197} & \underline{0.956} & 0.898 \\
		
		MSPred &  \textbf{3.41} & \textbf{0.812} & \underline{0.867} & \textbf{0.833} & \textbf{0.651} & \textbf{0.471} & 0.202 & 0.151 & 0.952 & \underline{0.901} \\
		
		\bottomrule
	\end{tabular}
	\vspace{-0.4cm}
\end{table}

%%%%%%%%%%%%%%%%%%%%%%%%%%%%%%%%%%%%%%%%%%%%%%%%%%%%%%%%%%%%%%%%%%%%%%%%%%%%%%%%
\vspace{-0.3cm}
\section*{Acknowledgments}
\vspace{-0.3cm}

This work was funded by grant BE 2556/16-2 (Research Unit FOR 2535 Anticipating
Human Behavior) of the German Research Foundation (DFG).

%\bibliography{egbib}
\bibliography{referencesAngel}

\setcounter{section}{0}
\renewcommand{\thesection}{\Alph{section}}

\vspace{-0.5cm}
\section{Datasets}
\label{section: datasets}

We evaluate MSPred for different prediction tasks on three video datasets of different levels of complexity, namely Moving MNIST~\cite{Srivastava_UnsupervisedLeaerningOfVideoRepresentationsUsingLSTMs_2015}, KTH-Actions~\cite{Schuldt_KTHRecognizingHumanActionsALocalSVMApproach_2004}, and SynpickVP.
\Table{table: datasets} summarizes the three datasets used in our work.

\textbf{Moving MNIST} is a standard video prediction dataset containing sequences of two random digits from the MNIST dataset~\cite{LeCun_MnistDataset_1998} moving with constant speed in a $64 \times 64$ grid, and bouncing off the image boundaries. In our experiments, we treat Moving MNIST frames as RGB images, i.e., repeating
the MNIST digits across the RGB channels. 
For the high-level representations, we use Gaussian blobs centered at the digit locations.
Despite its simplicity, this dataset is commonly used as a benchmark for video prediction.
For training, we randomly generate sequences of 49 frames by sampling two random MNIST digits, a starting position and speed; whereas for testing we use a fixed set containing 10,000 sequences.

\textbf{KTH-Actions} is a dataset consisting of real videos of humans performing one out of six possible actions, namely boxing, hand-clapping, hand-waving, walking, running and jogging. The dataset includes 600 videos of 25 different human actors performing the actions in various indoor and outdoor environments.
In our experiments, we downsample the images to a resolution of $64 \times 64$.
We use 1436 training sequences of length 49 from 16 different actors, whereas for testing we use 824 sequences from the remaining nine actors.
In addition to video frames, we use nine human keypoints as intermediate level representations, and a center-point of the person a high-level representation.
We generate the ground-truth keypoints using a pretrained OpenPose~\cite{Cao_RealTimePoseEstimationAffinityFields_2017} model for
human pose estimation.

\textbf{SynpickVP} is a new synthetic video prediction dataset, consisting of videos of various bin-picking scenarios in which a suction-cap gripper robot moves in arbitrary directions in a box containing different objects.
We generate the dataset by selecting sequences from the recently proposed SynPick~\cite{Periyasamy_SynPick_2021} dataset.
We use 1975 training and 200 evaluation sequences containing 29 RGB video frames of size $64 \times 112$.
This is a challenging video prediction benchmark, since the model needs to capture the motion of the
robotic gripper, as well as predict the future arrangement of displaced objects,
while still representing a complex and cluttered background.
In our experiments on SynpickVP, we train our model to predict image frames at the lowest level in the hierarchy, semantic segmentation maps from the 22 different classes at the intermediate level, and a single-keypoint heatmap for the robotic gripper position at the highest level.
Due to the synthetic nature of the dataset, semantic segmentation and object localization annotations are readily available.
When evaluating semantic segmentation forecasting, we average the class-wise results into three different categories: \emph{gripper} corresponds to the robot gripper, \emph{static} includes the different objects contained in the box, and \emph{background} corresponds to the red box where objects are placed.

\begin{table}[t!]
	\centering
	\caption{Summary of the datasets used in our experiments, including the size of the dataset splits and the type of high-level representations used for each dataset.}
	\label{table: datasets}
	\vspace{0.02cm}
	\small
	\begin{tabular}{p{2.7cm} wc{1.cm} wc{0.99cm} wc{0.99cm} wc{2.5cm} wc{2.1cm}}
		\toprule
		\textbf{Dataset Name} & \textbf{Img. Size} & \textbf{\# Train} & \textbf{\# Test} & \textbf{Mid-Level Rep.} & \textbf{High-Level Rep.} \\
		\midrule
		
		Moving MNIST~\cite{Srivastava_UnsupervisedLeaerningOfVideoRepresentationsUsingLSTMs_2015} & (3, 64, 64) & - & 10.000 & Digit Blob & Digit Position	\\
		
		KTH-Actions~\cite{Schuldt_KTHRecognizingHumanActionsALocalSVMApproach_2004} & (3, 64, 64) & 1.436 & 824 & Human Pose & Person Position	\\
		
		SynpickVP~\cite{Periyasamy_SynPick_2021} & (3, 64, 112) & 1.975 & 200 & Segmentation Maps & Gripper Position	\\
		
		\bottomrule
	\end{tabular}
\end{table}

\section{Evaluation Metrics}
\label{section: evaluation metrics}

We employ several evaluation metrics designed for different tasks 
in order to evaluate the predictions from the different MSPred decoder heads.
For future frame prediction, we compute several popular metrics which measure the visual
similarity between the predicted and ground-truth video frames. 
Furthermore, we employ different metrics to evaluate the ability of our model to make high-level structured predictions, such as human poses or semantic segmentation maps.
For all metrics, we average the results across all predicted frames or high-level structured representations.

\paragraph{Image Similarity Metrics:}  We evaluate our models for future frame
prediction using four popular metrics, namely MSE, PSNR, SSIM~\cite{Wang_SSIM_2004}, 
and LPIPS~\cite{Zhang_TheUnreasonableEffectivenessOfDeepFeaturesLPIPS_2018}.
MSE, PSNR and SSIM measure pixel or statistical differences between predicted and target images. However, they have been proven to correlate poorly with human perception,
favoring blurred predictions over more detailed, though imperfect, generations~\cite{Zhang_TheUnreasonableEffectivenessOfDeepFeaturesLPIPS_2018, Sara_ImageQualityAssesmentThroughSSIMMSEPSNTStudy_2019}.
Therefore, we favor LPIPS in our experiments, which measures the distance between CNN feature maps, and has been shown to better correlate with human judgment.

\vspace*{-0cm}
\paragraph{Pose and Keypoint Prediction Metrics:} MSPred forecasts future human poses at its intermediate level on the KTH-Actions dataset.
Given a predicted heatmap representing the location of a body joint, we extract the position coordinates by taking the location with maximum value of the heatmap, provided that the maximum value exceeds a certain threshold. Through empirical validation, we set the threshold value to 0.05.

In order to assess our model’s performance for human pose forecasting, we employ three popular metrics.
\emph{Mean Per Joint Position Error} (MPJPE) calculates the average $\ell$2-norm across predicted and target joints.
\emph{Percentage of Detected Joints} (PDJ) measures the fraction of the correctly estimated joints among the joints present in the ground-truth pose. A predicted keypoint is marked as a correct detection if its distance from the respective target keypoint does not exceed a certain threshold. We select this threshold as 20\% of the ground-truth person's height~\cite{Sapp_MultimodalDecomposableModelsForHumanPoseEstimation_2013}.
Similarly, \emph{Percentage of Correct Keypoints} (PCK) measures the fraction of correctly detected joints among the overall predicted joints.
Additionally, we also compute summary statistics for the PCK metric over
a range of thresholds. We evaluate the \emph{Average Precision} (AP)
as the mean PCK values computed over a range of thresholds {0.1, 0.2, ..., 0.5}.

\vspace*{-0cm}
\paragraph{Segmentation Metrics:} We predict semantic segmentation maps as the intermediate-level representation on the SynpickVP dataset.
We evaluate our predicted segmentation maps using two popular evaluation metrics.
\emph{Pixel accuracy} (Acc) measures the fraction of correctly classified pixels in the image, whereas \emph{Intersection over Union} (IoU) is computed by dividing the corresponding number of correctly estimated pixels, i.e. the area of overlap between predicted and ground-truth segments, by the area of union of the very segments.

We compute the average Acc and IoU metrics for three subsets of the classes. More precisely, we average the metrics separately for three object categories: robot gripper, static objects placed on the box, and the red box itself.

\section{Implementation Details}
\label{section: implementation details}

In this section we provide further implementation details of MSPred (\Section{subsection: mspred architecture}), and the hyper-parameter values used in our experiments (\Section{subsection: hyper-params}). Additionally we discuss the implementation of the SVG' baseline (\Section{subsection: SVG'}).
Our codebase is implemented using the PyTorch~\cite{Paszke_AutomaticDifferneciationInPytorch_2017} deep learning framework.
We run our experiments on an NVIDIA A6000 GPU with 48 GiB RAM.

\begin{table}[t!]
	%\caption{Encoder CNN architectures used for the Moving MNIST (left), and KTH-Actions and SynpickVP (right) experiments.}
	%\label{table: encoders}
	\vspace{0.1cm}
	\small
	\begin{minipage}[t]{.5\linewidth}
		\caption{DCGAN Discriminator Encoder}
		\label{table: DCGAN}
		\vspace{0.05cm}
		\centering
		\begin{tabular}{P{1.6cm} P{0.4cm} P{1.4cm} P{1.4cm}}
			\toprule
			\textbf{Layer} & \textbf{Size} & \textbf{Activation} & \textbf{Comment} \\
			\midrule
			Conv 4x4 & 64 & LeakyReLU & Stride 2 \\
			Conv 4x4 & 128 & LeakyReLU & Stride 2 \\
			Conv 4x4 & 256 & LeakyReLU & Stride 2 \\
			Conv 4x4 & 512 & LeakyReLU & - \\
			\bottomrule
		\end{tabular}
	\end{minipage}
	\begin{minipage}[t]{.5\linewidth}
		\caption{VGG16-Like Encoder}
		\label{table: VGG}
		\vspace{0.05cm}
		\centering
		\begin{tabular}{P{1.9cm} P{0.5cm} P{1.6cm}}
			\toprule
			\textbf{Layer} & \textbf{Size} & \textbf{Activation} \\
			\midrule
			2x Conv 3x3  & 64 & LeakyReLU  \\
			MaxPool 2x2 & - & - \\
			2x Conv 3x3  & 128 & LeakyReLU  \\
			MaxPool 2x2 & - & - \\
			2x Conv 3x3  & 256 & LeakyReLU  \\
			MaxPool 2x2 & - & -  \\
			2x Conv 3x3  & 512 & LeakyReLU \\
			\bottomrule
		\end{tabular}
	\end{minipage}
\end{table}

\subsection{MSPred Architecture Details}
\label{subsection: mspred architecture}

\paragraph{Encoder:} In order to ensure a fair comparison with baseline methods, the encoder is implemented following the SVG~\cite{Denton_StochasticVideoGenerationWithALearnedPrior_2018} architecture.
For the Moving MNIST dataset, the encoder follows the DCGAN discriminator~\cite{Radford_DCGAN_2015} architecture, whereas for KTH-Actions and SynpickVP we employ VGG16-like~\cite{Zisserman_VGG_2014} modules.
The architectures of both encoders are depicted in \Tables{table: DCGAN}{table: VGG}, respectively. All convolutional layers use padding `SAME', include a bias weight, and are followed by batch normalization~\cite{Ioffe_BatchNormalization_2015}.

\paragraph{Decoder:} The decoder in MSPred is implemented as a mirrored version of the corresponding encoder. In the DCGAN-like decoder, feature maps are upsampled via transposed convolutions, whereas in the VGG-like decoder upsampling is achieved via nearest neighbor interpolation.
The higher-level decoder heads are each composed of two convolutional blocks with the same structure as the decoder.

\paragraph{Predictor:} Our predictor module uses four ConvLSTM~\cite{Shi_ConvLSTMNetworkPrecipitationNowcasting_2015} cells for each of the three levels in the hierarchy, each with
128 kernels of size $3 \times 3$. The lowest-level LSTM processes all inputs, whereas the higher-level LSTMs process one out of every $\Period_1$ and $\Period_2$ inputs respectively.

\paragraph{Stochastic Component:} The prior ($LSTM_{\psi}$) and posterior ($LSTM_{\phi}$) modules are implemented as a single-cell ConvLSTM  with 64 kernels of size $3 \times 3$, followed by a convolutional layer mapping the feature maps into the desired latent dimensionality. Inspired by SVG~\cite{Villegas_HighFidelityVideoPrediction_2019}, we sample latent tensors with 10, 24 and 32 channels for the Moving MNIST, KTH-Actions, and SypickVP datasets, respectively.

\begin{table}[t!]
	\caption{Hyper-parameter values used for each dataset in our experiments}
	\label{table: hyper params}
	\centering
	\small
	\vspace{0.05cm}
	\begin{tabular}{P{2.4cm} P{0.01cm} P{2.2cm}P{2.0cm}P{1.6cm}}
		\toprule
		%\textbf{Hyper-Parameter} && \multicolumn{3}{c}{\textbf{Value}} \\
		\textbf{Hyper-Parameter} &&  \textbf{Moving MNIST} & \textbf{KTH-Actions} & \textbf{SynpickVP} \\
		\midrule
		$\Context$ && 9 & 9 & 9 \\
		$\Period_1$ && 4 & 4 & 2 \\
		$\Period_2$ && 8 & 8 & 4 \\
		Learning rate && $10^{-4}$ & $3 \cdot 10^{-4}$ & $5 \cdot 10^{-4}$ \\
		Batch size && 16 & 12 & 12 \\
		Num. Epochs && 350 & 800 & 200 \\
		$\lambda_1$ && 2.5 & 1.4 & 2.0 \\
		$\lambda_2$ && 2.5 & 0.2 & 0.3 \\
		$\beta$ && $10^{-4}$ & $5 \cdot 10^{-5}$ & $10^{-4}$ \\	
		\bottomrule
	\end{tabular}
\end{table}

\subsection{Hyper-Parameters}
\label{subsection: hyper-params}

The hyper-parameters used in our experiments are reported in \Table{table: hyper params}. We report the specific values for the experiments on each of the datasets.

\subsection{SVG'}
\label{subsection: SVG'}
As described in Section 4.4 of the paper, we train a specialized baseline \emph{SVG’}, based on a modified SVG-LP~\cite{Denton_StochasticVideoGenerationWithALearnedPrior_2018} model, which predicts high-level representations (e.g. human poses or semantic segmentation) conditioned on input video frames.
SVG' follows the same architecture as SVG-LP, but we apply some modifications to adapt the model for the tasks of pose and semantic segmentation forecasting, and for a fair comparison with MSPred.
First, the linear LSTM recurrent blocks are replaced by ConvLSTMs operating with a period of $\Period_1$, i.e., processing every $\Period_1$-th input.
Second, the number of output channels is changed from three to nine for KTH-Actions, and to 22 for SynpickVP. 
Finally, since there are no predicted image frames to be fed back into the model, we design SVG' to be autoregressive in the feature space, i.e., the output of the predictor module becomes its input at the subsequent time step.

\begin{figure}[t!]
	\begin{tabular}{ccc}
		\begin{minipage}{0.33\textwidth}
			\hspace{-0.43cm} \includegraphics[width=1\linewidth]{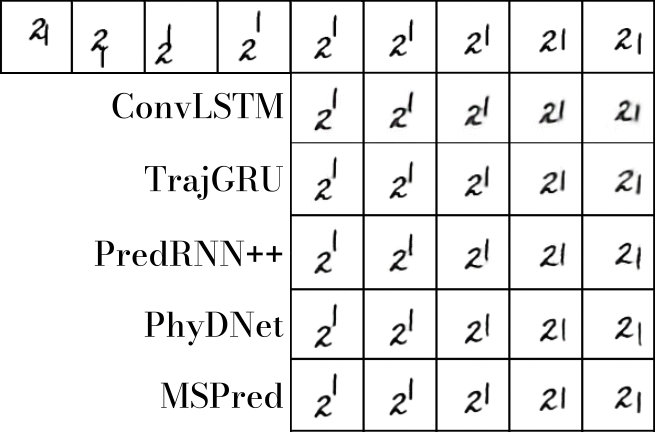} \\
		\end{minipage}
		&
		
		\begin{minipage}{0.33\textwidth}
			\vspace{-0.42cm}
			\hspace{-0.81cm} \includegraphics[width=1\linewidth]{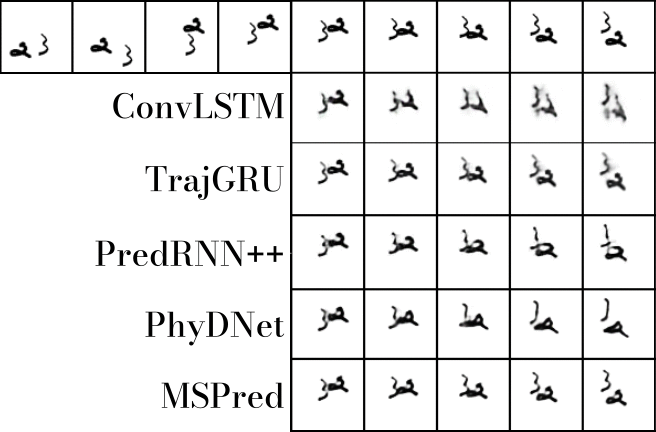}
		\end{minipage}	
		&
		\begin{minipage}{0.33\textwidth}
			\vspace{-0.43cm}
			\hspace{-1.19cm} \includegraphics[width=1\linewidth]{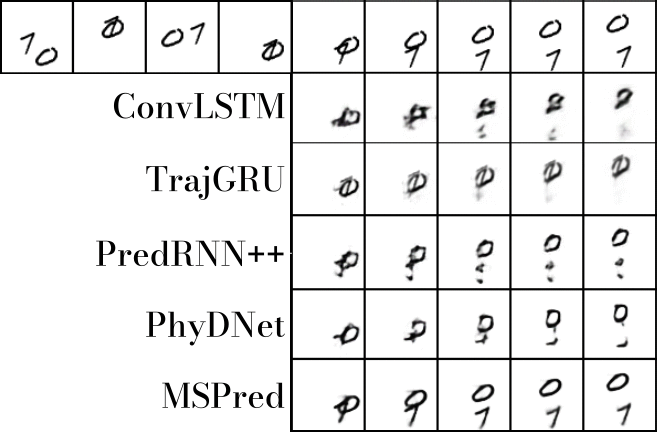}
		\end{minipage}	
	\end{tabular}
	\vspace{-0.56cm}
	\caption{
		Qualitative results on the Moving MNIST dataset. 
		Top row corresponds to ground truth frames. 
		We display four seed frames and five predictions for three test-set sequences.
		In general, all compared methods achieve good frame
		predictions. However, only MSPred accurately resolves challenging cases in which digits overlap.
		Colors are inverted to improve the visualization.
	}
	\label{fig:mmnist qual}
\end{figure}

\begin{figure}[t!]
	\begin{tabular}{ccc}
		\begin{minipage}{0.33\textwidth}
			\hspace{-0.43cm} \includegraphics[width=1\linewidth]{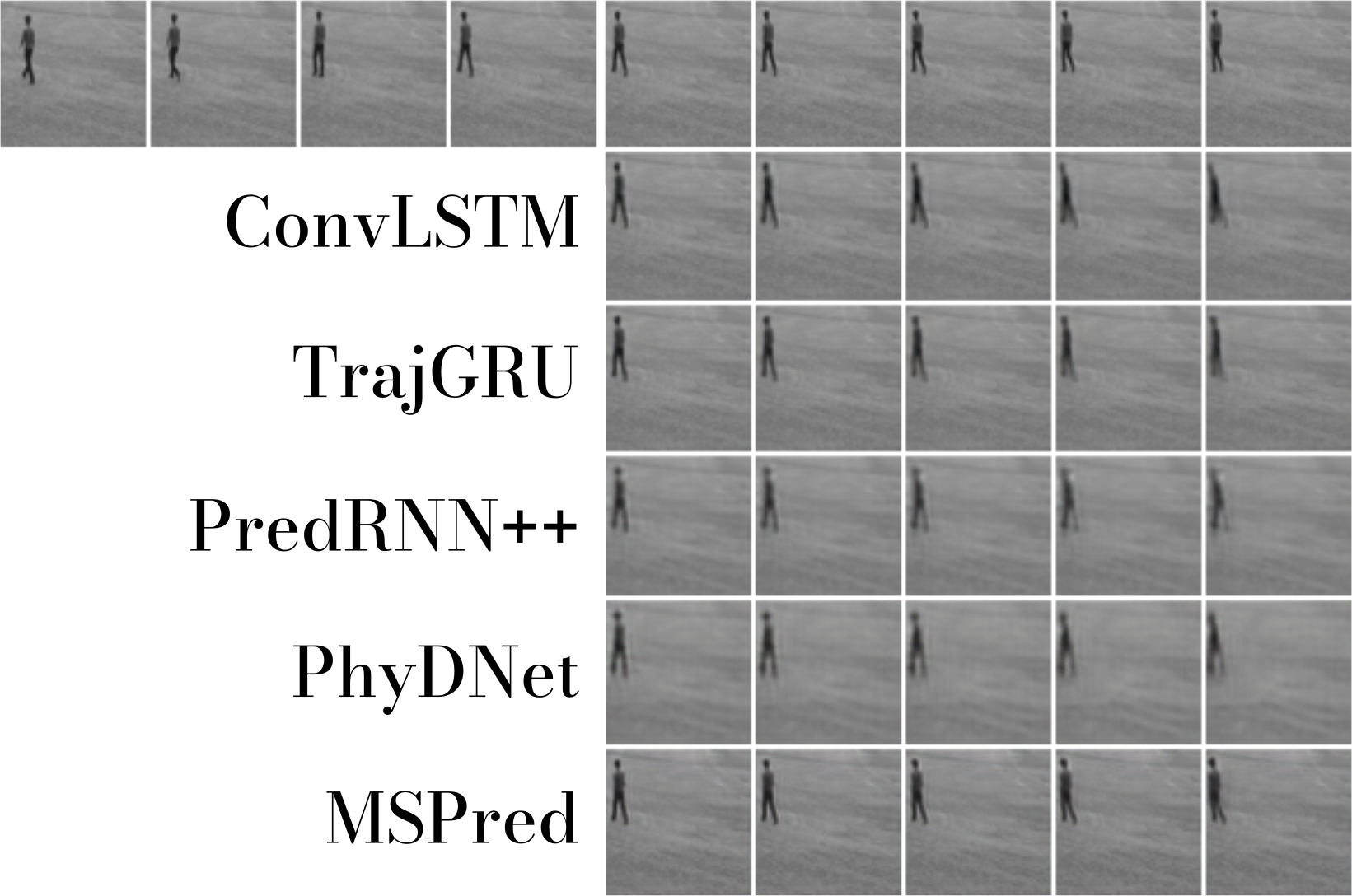} \\
		\end{minipage}
		&
		\begin{minipage}{0.33\textwidth}
			\vspace{-0.41cm}
			\hspace{-0.81cm} \includegraphics[width=1\linewidth]{imgs/kth_2.png}
		\end{minipage}	
		&
		\begin{minipage}{0.33\textwidth}
			\vspace{-0.41cm}
			\hspace{-1.2cm} \includegraphics[width=1\linewidth]{imgs/kth_3.png}
		\end{minipage}	
	\end{tabular}
	\vspace{-0.7cm}
	\caption{
		Qualitative results on the KTH-Actions dataset. 
		Top row corresponds to ground truth frames. 
		We display four seed frames and five predictions for three test-set sequences.
		MSPred achieves the sharpest and more accurate predictions among the compared methods.
	}
	\label{fig:kth qual}
\end{figure}

\begin{figure}[t!]
	\vspace{-0.1cm}
	\begin{tabular}{ccc}
		\begin{minipage}{0.495\textwidth}
			\hspace{-0.45cm} \includegraphics[width=1\linewidth]{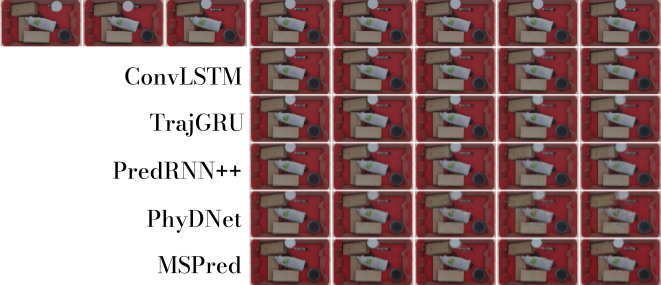} \\
		\end{minipage}
		&
		\begin{minipage}{0.495\textwidth}
			\vspace{-0.45cm}
			\hspace{-0.8cm} \includegraphics[width=1\linewidth]{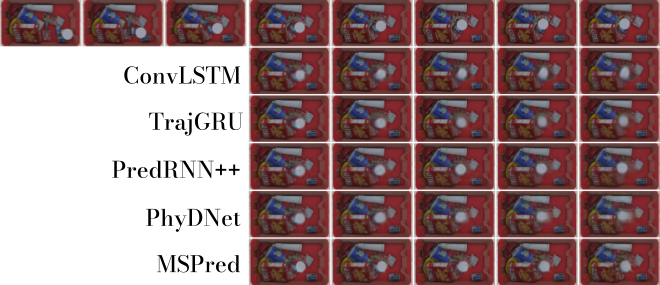}
		\end{minipage}	
	\end{tabular}
	\vspace{-0.7cm}
	\caption{
		Qualitative results on the SynpickVP dataset. 
		Top row corresponds to ground truth frames. 
		We display three seed frames and five predictions for two test-set sequences.
		MSPred qualitatively outperforms the compared methods, achieving sharp reconstructions, whereas the baseline methods tend to blur the predictions.
	}
	\label{fig:synpick qual}
\end{figure}

\section{Qualitative Results}
\label{section: qualitative results}

\begin{figure}[t]
	\begin{tabular}{cc}
		\begin{minipage}{0.495\textwidth}
			\hspace{-0.45cm} \includegraphics[width=1\linewidth]{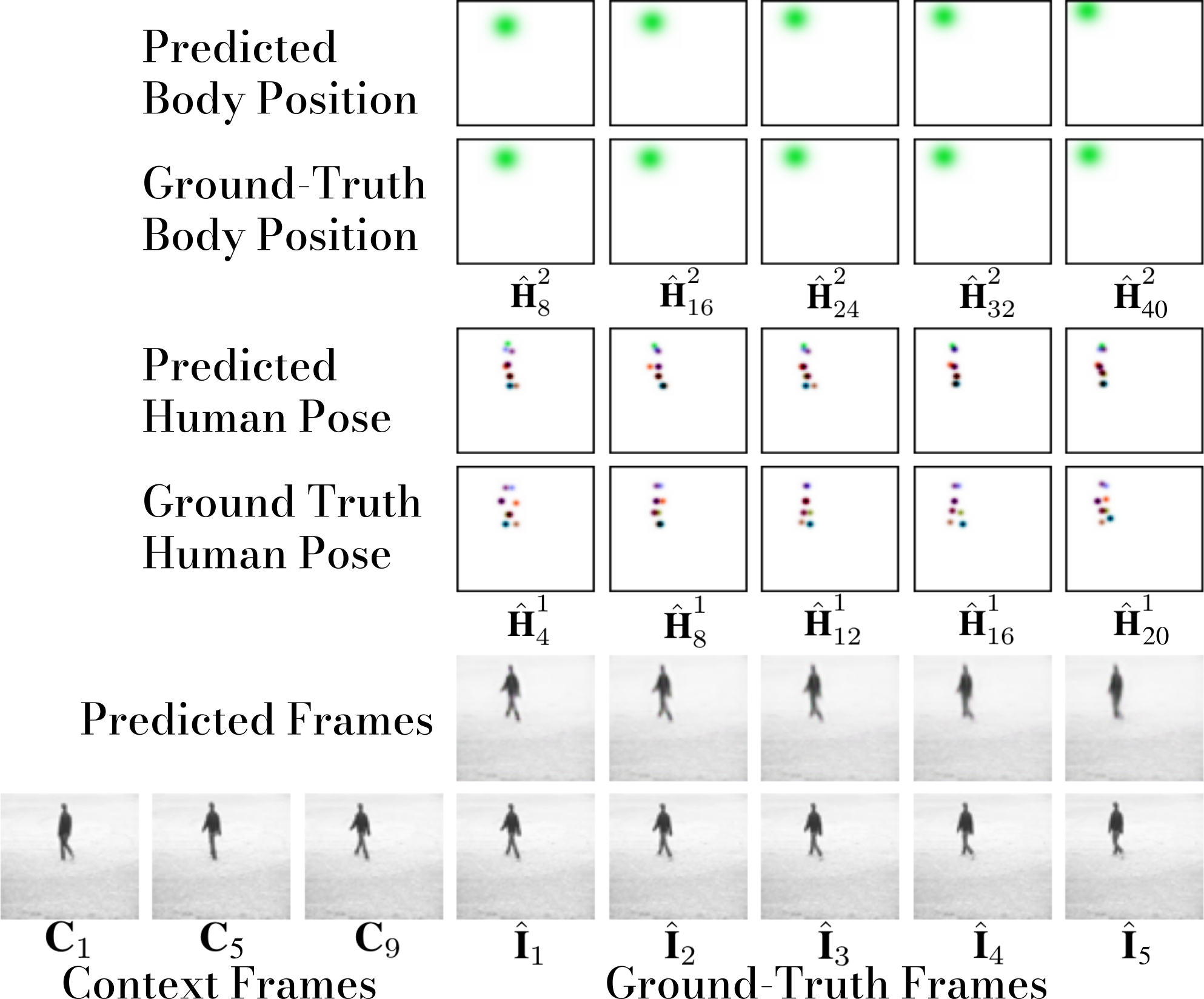} \\
		\end{minipage}
		&
		\begin{minipage}{0.495\textwidth}
			\vspace{-0.38cm}
			\hspace{-0.8cm} \includegraphics[width=1\linewidth]{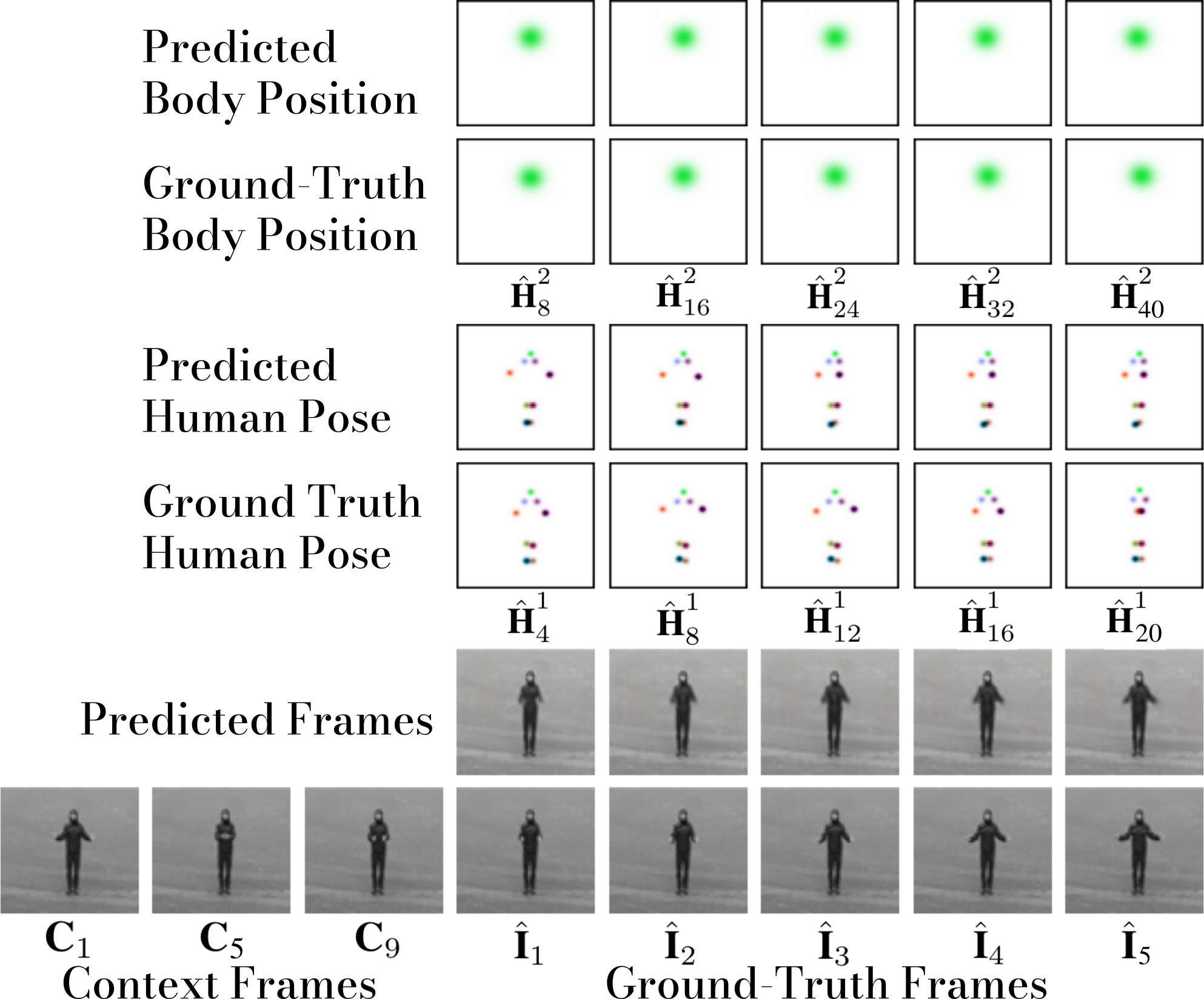}
		\end{minipage}	
	\end{tabular}
	\vspace{-0.5cm}
	\caption{
		Predictions of different level of abstraction on the KTH-Actions dataset. We display three seed frames and five targets and predictions for each decoder head. 
		MSPred forecasts frames on short time horizons, while also predicting human poses and person locations longer into the future using coarser temporal resolutions.
	}
	\label{fig: ms kth}
\end{figure}

In \Figuress{fig:mmnist qual}{fig:synpick qual 2}, we qualitatively compare several video prediction models for the task of future frame prediction on the Moving MNIST, KTH-Actions and SynpickVP datasets, respectively.
\Figures{fig: ms kth}{fig: ms synp} depict additional examples of multi-scale prediction on the KTH and SynpickVP datasets, respectively.
Further images and animations can be found in the project website\footnote{https://sites.google.com/view/mspred/home}.

\begin{figure}[t!]
	\vspace{-0.1cm}
	\begin{tabular}{ccc}
		\begin{minipage}{0.495\textwidth}
			\hspace{-0.45cm} \includegraphics[width=1\linewidth]{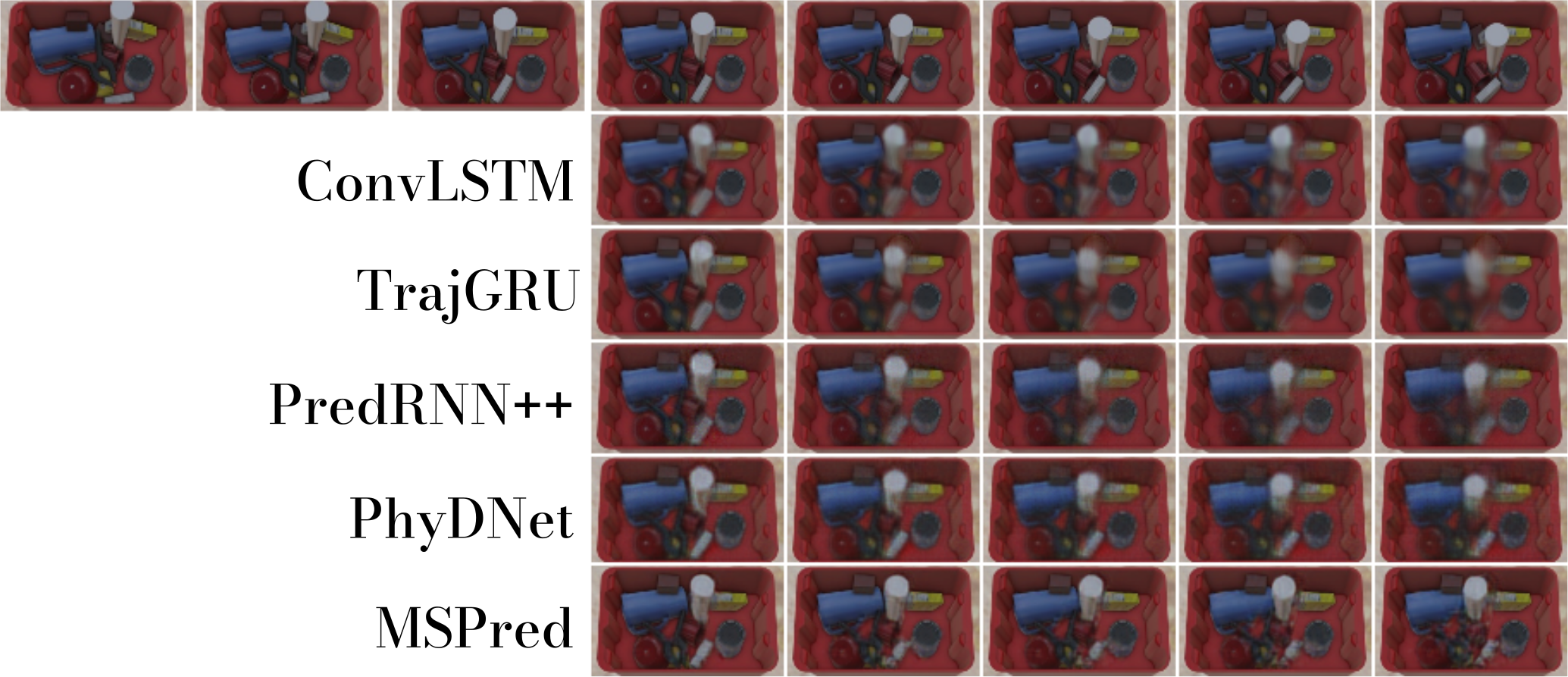} \\
		\end{minipage}
		&
		\begin{minipage}{0.495\textwidth}
			\vspace{-0.45cm}
			\hspace{-0.8cm} \includegraphics[width=1\linewidth]{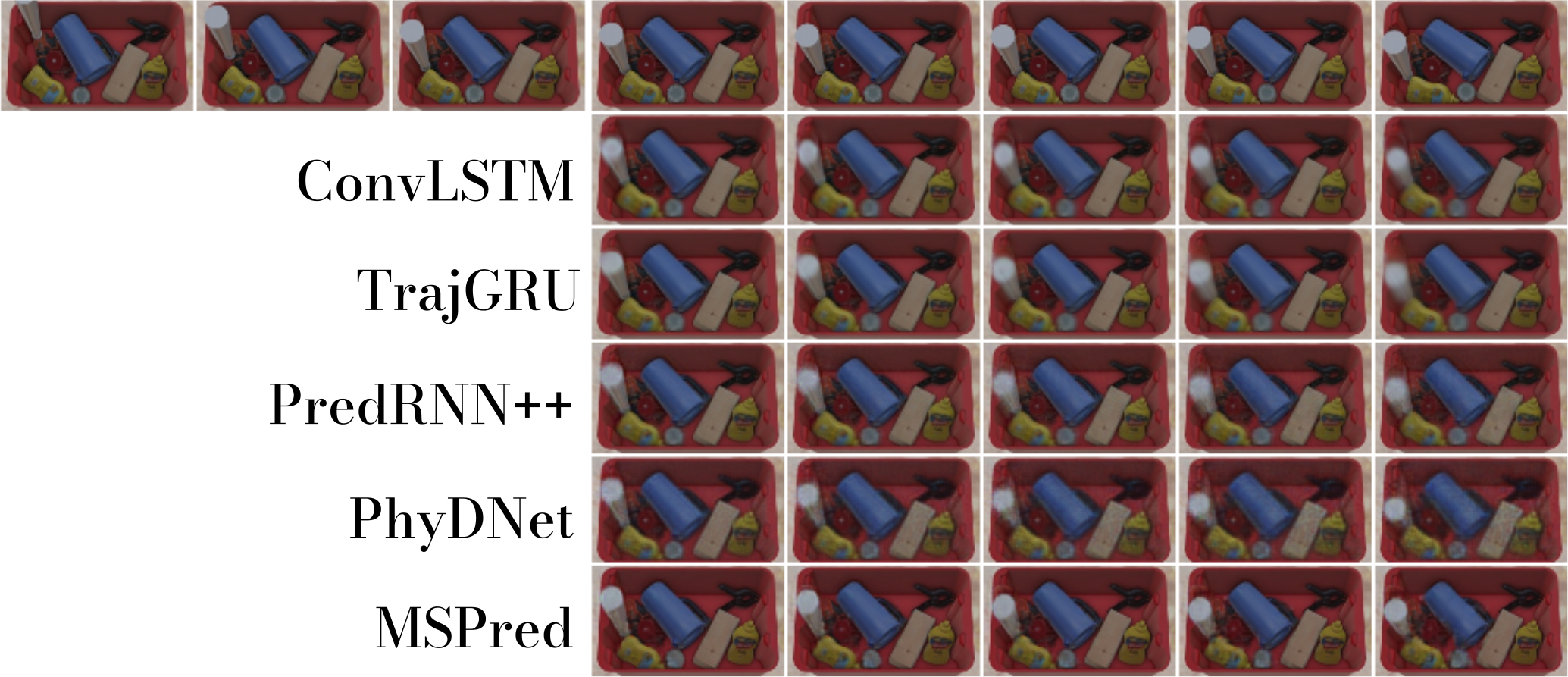}
		\end{minipage}	
	\end{tabular}
	\vspace{-0.7cm}
	\caption{
		Qualitative results on the SynpickVP dataset. 
		Top row corresponds to ground truth frames. 
		We display three seed frames and five predictions for two test-set sequences.
		MSPred qualitatively outperforms the compared methods, achieving sharp reconstructions, whereas the baseline methods tend to blur the predictions.
	}
	\label{fig:synpick qual 2}
\end{figure}

\begin{figure}[t]
	\begin{tabular}{cc}
		\begin{minipage}{0.495\textwidth}
			\hspace{-0.4cm} \includegraphics[width=1\linewidth]{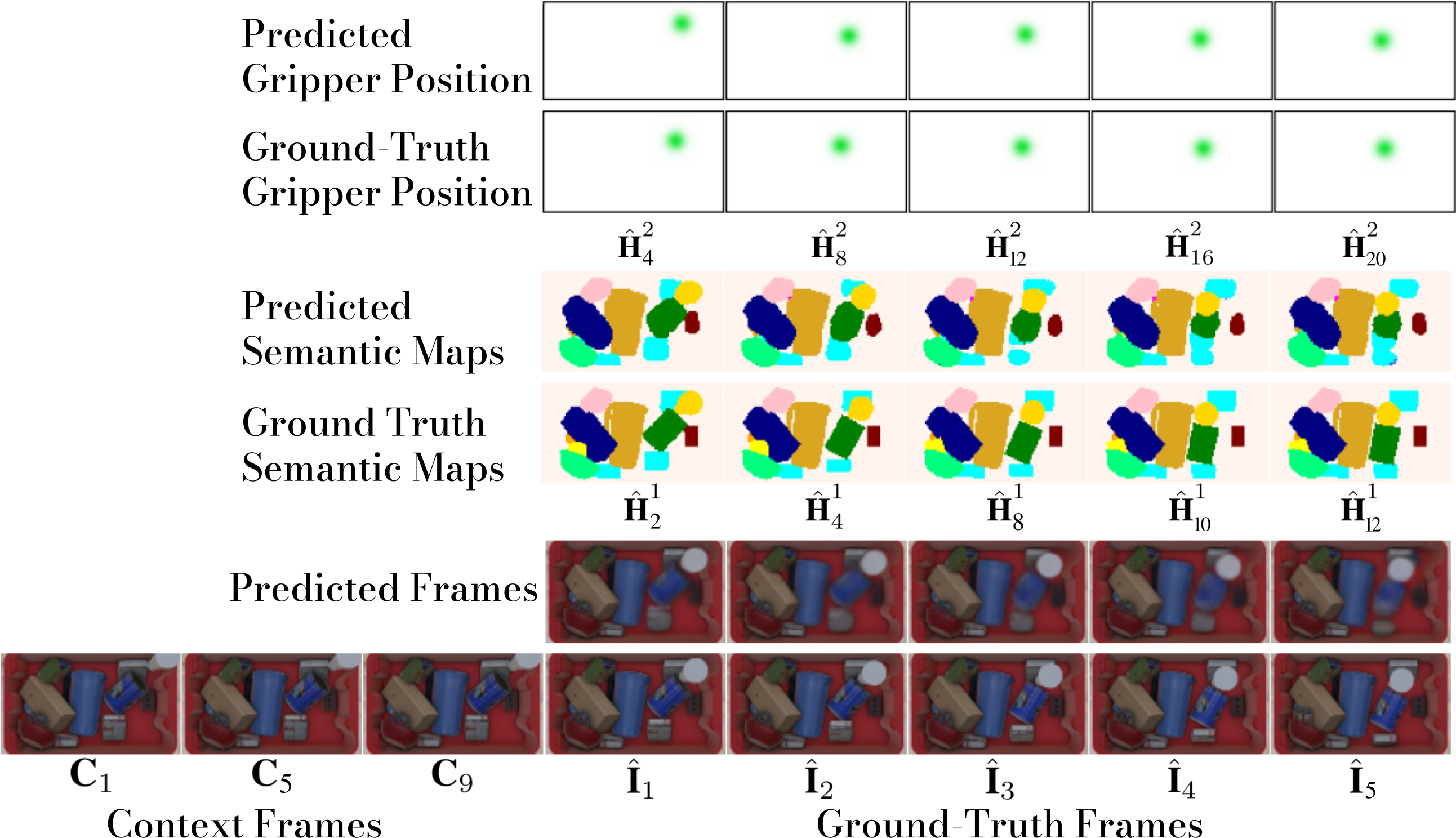} \\
		\end{minipage}
		&
		\begin{minipage}{0.495\textwidth}
			\vspace{-0.42cm}
			\hspace{-0.8cm} \includegraphics[width=1\linewidth]{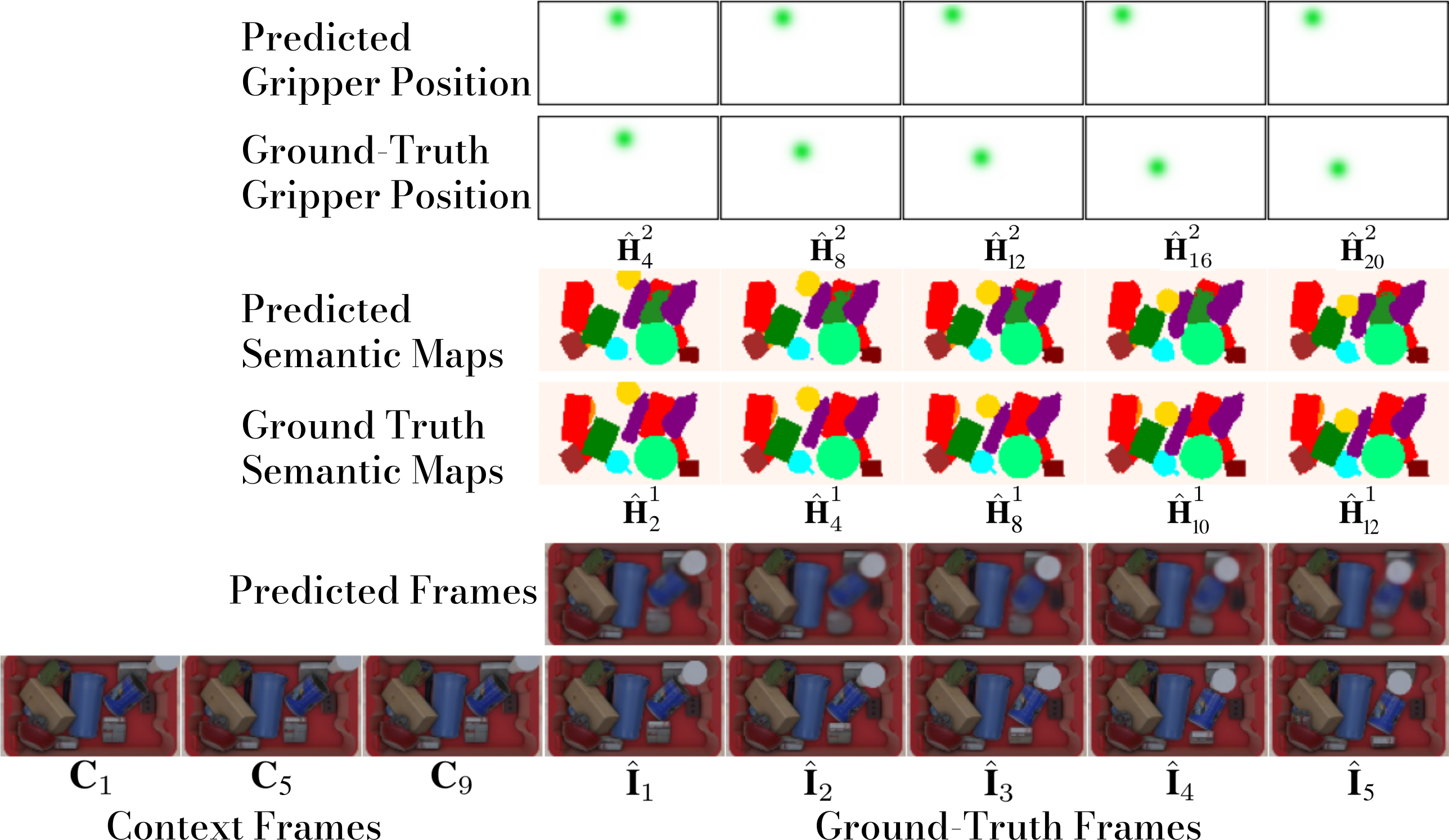}
		\end{minipage}	
	\end{tabular}
	\vspace{-0.5cm}
	\caption{
		Predictions of different levels of abstraction on the SynpickVP dataset. We display three seed frames, and five targets and predictions for each decoder head. 
		MSPred forecasts frames on short time horizons, while also predicting the semantic segmentation of the scene and the gripper location long into the future using coarser temporal resolutions.
	}
	\label{fig: ms synp}
\end{figure}

\end{document}